%% file: main_arxiv.tex
\documentclass[runningheads]{llncs}
\usepackage{graphicx}
\usepackage{todonotes}
\usepackage{standalone}
\usepackage{amsfonts}       
\usepackage{nicefrac}       
\usepackage{amsmath}
\usepackage{amssymb}
\usepackage{rotating}
\usepackage{booktabs} 
\usepackage{multirow}

\usepackage{hyperref}

\usepackage{stmaryrd}

\newcommand*{\defeq}{\mathrel{\vcenter{\baselineskip0.5ex \lineskiplimit0pt
			\hbox{\footnotesize.}\hbox{\footnotesize.}}}%
	=}

\newcommand{\selectors}{\mathcal{S}} 
\newcommand{\algorithms}{\mathcal{A}}

\usepackage{xcolor}

\makeatletter
\newcommand\footnoteref[1]{\protected@xdef\@thefnmark{\ref{#1}}\@footnotemark}
\makeatother
\begin{document}

\title{Algorithm Selection on a Meta Level}

\authorrunning{A. Tornede et al.}

\author{
    Alexander Tornede\inst{1} 
    \and
    Lukas Gehring\inst{1} 
    \and
    Tanja Tornede\inst{1}
    \and
    Marcel Wever\inst{1}   
    \and
    Eyke H{\"u}llermeier\inst{2} 
}



\institute{Institut f{\"u}r Informatik, Paderborn University, 33098 Paderborn, Germany 
\email{\{alexander.tornede@, lgehring@mail., tanja.tornede@, marcel.wever@\}upb.de}
\and
University of Munich (LMU), 80538 Munich, Germany \\
\email{eyke@ifi.lmu.de}}

\date{Received: date / Accepted: date}

\maketitle

\begin{abstract}
The problem of selecting an algorithm that appears most suitable for a specific instance of an algorithmic problem class, such as the Boolean satisfiability problem, is called instance-specific algorithm selection. Over the past decade, the problem has received considerable attention, resulting in a number of different methods for algorithm selection. Although most of these methods are based on machine learning, surprisingly little work has been done on meta learning, that is, on taking advantage of the complementarity of existing algorithm selection methods in order to combine them into a single superior algorithm selector. In this paper, we introduce the problem of meta algorithm selection, which essentially asks for the best way to combine a given set of algorithm selectors. We present a general methodological framework for meta algorithm selection as well as several concrete learning methods as instantiations of this framework, essentially combining ideas of meta learning and ensemble learning. In an extensive experimental evaluation, we demonstrate that ensembles of algorithm selectors can significantly outperform single algorithm selectors and have the potential to form the new state of the art in algorithm selection.

\end{abstract}

\section{Introduction}

Looking at algorithmic problem classes such as Boolean satisfiability (SAT) \cite{satzilla07_xu2007,satzilla11_xu2011hydra}, the traveling salesman problem (TSP) \cite{DBLP:conf/ictai/PiheraM14}, or constraint satisfaction (CSP) \cite{lobjois1998branch}, practical experience suggests that algorithms perform differently on different problem instances: while algorithm $A$ might be better than $B$ on a specific instance (e.g., a specific TSP), $B$ may outperform $A$ on another instance (e.g., another TSP). This is not very surprising and completely in line with theoretical results proving that there is ``no free lunch'', i.e., excluding that one algorithm uniformly dominantes all others \cite{no_free_lunch_wolpert1997no}.
The following task thus appears to be meaningful from a practical point of view: Given a problem class and a pool of algorithms to choose from, find a rule that automatically assigns a (presumably) most suitable algorithm to each possible problem instance. This task is called (instance-specific) algorithm selection (AS) in the literature \cite{rice1976algorithm}. Here, suitability may refer to different performance criteria, such as runtime \cite{tornede20_run2survive} or a measure of solution quality \cite{tpami}.

The problem of algorithm selection has received considerable attention over the past decade, resulting in a large set of heterogeneous algorithm selection approaches. Many of these approaches rely on machine learning, which essentially means that a rule assigning algorithms to problem instances is learned from suitable training data, for example, the performance observed in the past when running specific algorithms on specific instances. Given a new instance, a machine learning algorithm leverages such data to predict the performance of the candidate algorithms, or to predict the presumably best algorithm directly.  
AS approaches of that kind achieve state-of-the-art performance and typically outperform the best stand-alone algorithm, also referred to as ``single best solver'' (SBS) in the following, by several orders of magnitude \cite{kerschkeHNT19_as_survey}.

Interestingly, because an algorithm selector is again an algorithm (taking an instance as input and returning a presumably best algorithm as output), the very same task of algorithm selection can also be considered on a meta level, giving rise to the following question: Given a problem instance and a set of algorithm selectors, which one should be used to predict the best algorithm?  
This question could be answered by an algorithm selector on the meta level, that is, by an ``algorithm selector selector'', which does not choose among the algorithms (or ``base algorithms'', to distinguish them from the AS algorithms), but among the algorithm selectors, which in turn are responsible for selecting an algorithm. Indeed, a certain complementarity among AS approaches can be observed (e.g.~\cite{tornede20_run2survive}) and the resulting meta-AS problem was first mentioned by \cite{lindauerRK19_as_competitions} and \cite{kerschkeHNT19_as_survey}, though without pursuing it further. 

Having the choice between a set of candidate algorithm selectors, limiting oneself to choosing only \textit{a single} one of them (which in turn chooses the final algorithm) might actually seem unnecessarily restrictive. In fact, leveraging a \emph{composition} of selectors, which then choose the final algorithm jointly, might be a better idea. 
This naturally leads to \textit{ensemble learning} \cite{dietterich00_ensembles}, which is a common approach in machine learning to combine several predictors into stronger compositions. Thus, instead of using a single algorithm selector to choose an algorithm, a set of selectors is asked to evaluate the available algorithms. Subsequently, these evaluations are aggregated into a joint decision.
Somewhat surprisingly, building ensembles of algorithm selectors has hardly been considered in the AS literature so far (see \autoref{sec:related-work}), although ensemble learning is well known to improve predictive accuracy in standard machine learning problems such as classification and regression. One reason could be that querying multiple models obviously takes more time than querying only a single one, so that ensembling may appear counterintuitive in scenarios where runtime is considered as the target measure.

In this paper, we formalize the problem of meta algorithm selection and propose algorithmic solutions. Furthermore, we investigate their potential to make better decisions with respect to the selection of algorithms. In an extensive empirical study, we find that trying to learn the best algorithm selector, i.e., to predict which algorithm selector will pick the best algorithm for a given query, does not lead to better algorithm selection performance. On the other side, ensembling algorithm selectors helps to improve efficacy, while the additional runtime consumed for querying multiple algorithm selectors remains negligible.
Of course, the improved performance comes at a higher cost of building the ensemble algorithm selector, because multiple basic algorithm selectors need to be fitted for one ensemble.
However, this does not pose a problem in practice, because algorithm selectors are in general built in an offline phase prior to the actual selection process.

The remainder of the paper is structured as follows. First, we give a formal introduction to the algorithm selection problem in Section~\ref{sec:algorithm-selection}, followed by a definition of the meta AS problem in Section~\ref{sec:meta_as} and a first (still quite limited) solution to the problem in Section~\ref{sec:meta-learning_ass}. As a more advanced solution, we present algorithm selection ensembles in Section~\ref{sec:ensembles}. Subsequently, we present and discuss the results of our empirical evaluation in Section~\ref{sec:evaluation}. Related work is discussed in Section~\ref{sec:related-work}, prior to concluding our paper in Section~\ref{sec:conclusion}.

\section{Algorithm Selection}\label{sec:algorithm-selection}

In the per-instance algorithm selection problem, first formalized by \cite{rice1976algorithm}, we are faced with a space of instances $\mathcal{I}$ of an algorithmic problem class (such as SAT, where every instance is a logical formula) and a finite set of algorithms $\mathcal{A}$, which solve such instances. The goal is to find a map $s: \mathcal{I} \longrightarrow \mathcal{A}$, called algorithm selector, which assigns algorithms to instances. An assignment $a = s(i)$ is interpreted as a recommendation, suggesting that algorithm $a \in \mathcal{A}$ will perform strongly, or perhaps even best among all algorithms, on problem instance $i \in \mathcal{I}$. More formally, the goal is to maximize (expected) performance in terms of a measure $m: \mathcal{I} \times \mathcal{A} \longrightarrow \mathbb{R}$, which is also part of the AS problem specification. Hence, the optimal algorithm selector for all instances $i \in \mathcal{I}$, also known as the \textit{oracle} or \textit{virtual best solver} (VBS), is defined as 
\begin{equation}
    s^*(i) = \arg\min_{a \in \mathcal{A}} \mathbb{E}\left[ m(i,a) \right] \, ,
\end{equation}
where the expectation accounts for the potential randomness imposed by the algorithm.
We denote the algorithm that is best on average (in expectation) on a predefined set of instances as the \textit{single-best solver} (SBS). It constitutes the default baseline in algorithm selection. 

Observe that an exhaustive evaluation of all algorithms for computing the VBS is not deemed a solution, because $m$ is usually costly to evaluate and often even requires running the respective algorithm. For example, if runtime is the measure of interest, a single evaluation already results in a solved instance, rendering all other evaluations unnecessary. Hence, instead of performing evaluations at query time, the algorithm selector should make use of gathered knowledge to come to a decision.

\subsection{Algorithm Selection Methods}\label{subsec:as_solutions}

The majority of AS approaches leverages machine learning techniques to learn (in one way or another) a surrogate performance measure $\widehat{m}: \mathcal{I} \times \mathcal{A} \longrightarrow \mathbb{R}$ approximating $m$ while being cheap to evaluate. With such a surrogate performance measure at hand, an exhaustive enumeration, actually excluded for the reasons explained before, does become possible and yields the canonical algorithm selector
\begin{equation}\label{eq:canonical_as}
    s(i) \defeq \arg\min_{a \in \mathcal{A}} \widehat{m}(i,a) \,.
\end{equation}
For the purpose of inferring such a surrogate, the setting is usually assumed to contain a set of training instances $\mathcal{I}_D \subset \mathcal{I}$ on which some (but not necessarily all) of the algorithms in $\mathcal{A}$ have been evaluated, so that performance evaluations $m(i,a)$ are available. Note that the corresponding training performance matrix spanned by $\mathcal{I}_D$ and $\mathcal{A}$ is usually assumed to contain (sometimes many) missing values. Furthermore, instances are assumed to be representable by a set of $d$ features generated by a feature map $f: \mathcal{I} \longrightarrow \mathbb{R}^d$. In many cases, such features are available or can be defined in a quite natural way. In the case of SAT, for example, common features include the length of a formula, the number of clauses or variables, etc. In general, the computation of features does not come for free and requires time. This should be taken into account, especially when runtime is chosen as a performance measure to be optimized.

One of the most straight-forward instantiations of the framework described above, PerAlgo, was proposed by \cite{satzilla07_xu2007}, where one performance surrogate $\widehat{m}_a: \mathcal{I} \longrightarrow \mathbb{R}$ is learned for each algorithm $a \in \mathcal{A}$ separately. The joint surrogate can then be defined as $\widehat{m}(i,a) = \widehat{m}_a(i)$ for all instances $i \in \mathcal{I}$. 

Alternatively, the problem can be formalized as a multi-class classification problem, where each algorithm corresponds to a class, so that a multi-class classifier (Multiclass) of the form $s: \mathcal{I} \longrightarrow \mathcal{A}$ can be learned directly. A well-known example from this category is SATzilla'11 \cite{satzilla11_xu2011hydra}, which employs an all-pairs decomposition approach, learning a cost-sensitive classifier for each pair of algorithms and determining the selected algorithm by majority voting. Building upon the idea of pairwise comparisons of algorithms, \cite{crr_hanselle2020} suggest learning selectors via a combined ranking and regression approach. Similarly, \cite{kotthoff12} suggests employing a stacking approach, using regression models to predict the performance of each algorithm, which is used as an additional input for a meta-learner selecting the final algorithm.

Focusing on so-called censored information present in algorithm selection data, \cite{tornede20_run2survive} propose a decision-theoretic approach (R2S-PAR10 and R2S-EXP), leveraging techniques from survival analysis to effectively learn from such censored information. Similarly, \cite{pakdd_hanselle2021algorithm} consider the censored information present in the data within the framework of superset learning \cite{superset_learning_HULLERMEIER20141519}.

Furthermore, instance-based approaches, such as SUNNY \cite{sunny_amadiniGM14} or ISAC \cite{isac_kadiogluMST10}, have proven to successfully perform algorithm selection by exploiting performances recorded on similar instances in the training data. To this end, they employ k-nearest neighbor or clustering techniques in order to estimate the performance of an algorithm on an unseen instance. 

Finally, \cite{extreme_algorithm_selection_tornedeWH20} and \cite{tornede2019algorithm} propose the setting of ``extreme algorithm selection'', in which the pool of algorithms to choose from can be extremely large. They show that, by leveraging a feature representation not only for problem instances but also for algorithms, convincing selection performance can be achieved even in this setting.

\subsection{Loss Functions}

\begin{figure}[t]
    \centering
    \includegraphics[width=0.75\textwidth]{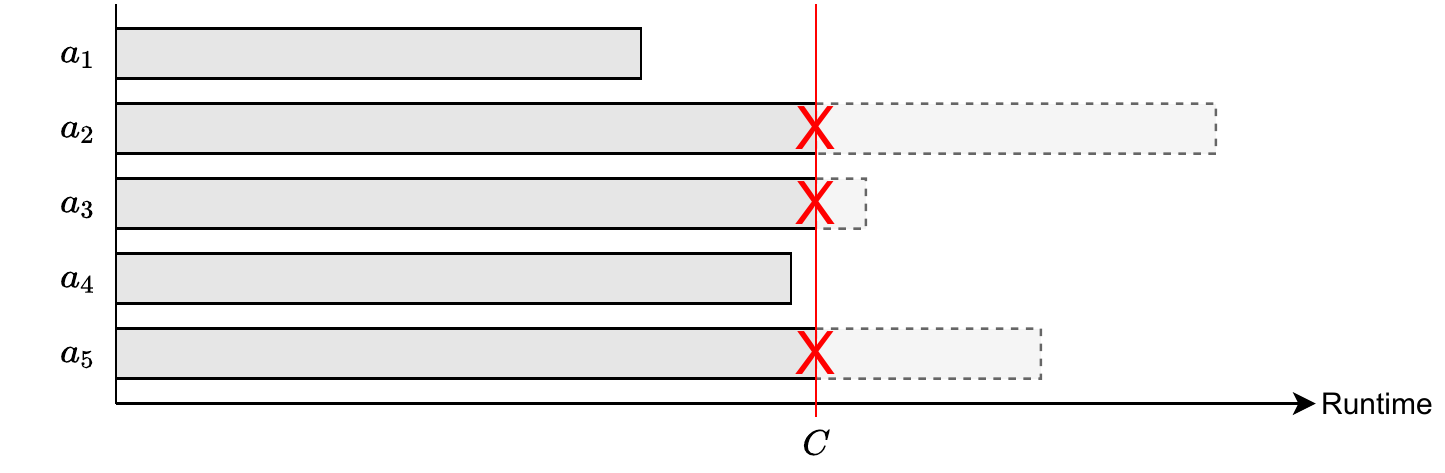}
    \caption{This figure depicts the process of running multiple algorithms on an instance (e.g. for training data generation). If an algorithm requires longer than $C$ to solve an instance, it is forcefully terminated and a selection of the corresponding algorithm will be punished.}
    \label{fig:algorithm_cutoff}
\end{figure}

One of the most natural and interesting performance measures to consider for satisfaction problems is the time until the instance is solved, i.e., the algorithm runtime. Unfortunately, combinatorial problems often feature skewed runtime distributions, such that some algorithms are running extremely long on some instances \cite{gomesSC97_heavy_tails}. As a consequence, algorithms are generally executed with an upper bound $C$ on their runtime. If an algorithm does not terminate within this bound, called \textit{cutoff}, the instance is considered unsolved and the algorithm is forcefully terminated; see \autoref{fig:algorithm_cutoff} for an illustration. As choosing an algorithm running into a cutoff leads to an unsolved instance, such a choice should be avoided by all means. One of the most common loss functions in AS, called the penalized average runtime (\textit{PAR10}), considers this by explicitly penalizing such timeouts. The \textit{PAR10} over a set of instances $\mathcal{I}' \subset \mathcal{I}$, called scenario, is defined as follows, where $m(i,s(i))$ corresponds to the runtime of the algorithm $s(i)$ chosen by the algorithm selector $s$ (and potentially the time required to compute the corresponding instance features) on instance $i$:
\begin{equation}\label{eq:par10}
\begin{split}
\mathit{PAR10}(s, \mathcal{I}') &= \frac{1}{\vert \mathcal{I}' \vert} \sum\limits_{i \in \mathcal{I}'} PR10(s,i) \\
\mathit{PR10}(s, i) &=
\begin{cases}
    m(i,s(i)) & \text{if } m(i,s(i)) \leq C \\
    10 \cdot C & \text{else}
\end{cases} 
\end{split}  
\end{equation} 
Naturally, \textit{PAR10} scores can vary drastically across scenarios making them incomparable. To alleviate this situation, one often falls back to the normalized PAR10 score of an algorithm selector $s$ defined as
\begin{equation}\label{eq:npar10}
    \mathit{nPAR10}(s, \mathcal{I}') = \frac{\mathit{PAR10}(s, \mathcal{I}') - \mathit{PAR10}(\mathit{oracle}, \mathcal{I}')}{\mathit{PAR10}(\mathit{SBS},\mathcal{I}') - \mathit{PAR10}(\mathit{oracle},\mathcal{I}')} \,.
\end{equation}
An \textit{nPAR10} score of $0$ corresponds to the oracle performance, a score of $1$ corresponds to a performance on a par with the SBS, whereas scores above $1$ indicate a deterioration in comparison to the SBS. Therefore, lower \textit{nPAR10} scores indicate better performance, and a successful algorithm selector should definitely have a score of less than $1$.

\section{Meta Algorithm Selection}\label{sec:meta_as}

Similar to the algorithms actually solving the problem instances, the algorithm selectors also show the phenomenon of performance complimentarity, as mentioned earlier. This gives rise to the question whether choosing between different algorithm selectors might be beneficial. In fact, by moving to the meta level, i.e. from the level of choosing among algorithms to the level of choosing among the algorithm selectors, we gain more freedom and can even select \textit{multiple selectors} instead of only a single algorithm as long as we ensure to aggregate the selections made by the selectors such that a single algorithm is returned at the end. Thus, the problem of per-instance meta algorithm selection (meta AS) concerns the problem of selecting one or multiple algorithm selectors together with an aggregation, for a given instance of an algorithmic problem class.
Each of the selected algorithm selectors then in turn selects an algorithm for solving the problem. Finally, these selected algorithms are aggregated such that only a single algorithm (of these) is returned. Hence, instead of directly choosing an algorithm to solve a problem instance, we take a detour by selecting one or multiple algorithm selectors and aggregating their decisions.

Formally, in the meta AS problem, we are given a set of algorithm selectors $\selectors \subseteq \{ s \vert s: \mathcal{I} \longrightarrow \algorithms \}$, which is a subset of all possible selection functions, in addition to the instance space $\mathcal{I}$, the set of algorithms $\mathcal{A}$ and the performance measure $m$ known from the AS problem. We then seek to find a mapping 
\begin{equation}\label{eq:ass}
   \mathit{ass}: \mathcal{I} \longrightarrow 2^{\selectors} \,,
\end{equation}
called algorithm selector selector (ASS), and an aggregation function
\begin{equation}\label{eq:agg}
    \mathit{agg}: \mathcal{I} \times 2^{\selectors} \longrightarrow \algorithms \, ,
\end{equation} 
such that the algorithm resulting from the aggregation optimizes the original performance measure $m$. Accordingly, we seek to find the best pair $(\mathit{agg}, \mathit{ass})$ of aggregation function $\mathit{agg}$ and algorithm selector selector $\mathit{ass}$, such that for all instances $i \in \mathcal{I}$ the best algorithm is returned, i.e., 
\begin{equation}
    \mathit{agg}(i, \mathit{ass}(i)) \in \arg\min_{a \in \algorithms} \mathbb{E}\left[m(i,a)\right] \,\,\,.
\end{equation}
Observe that we principally allow the concrete aggregation to depend on the instance, thereby allowing for \textit{learning instance-specific aggregation} functions.

\begin{figure}[t]
    \centering
    \includegraphics[width=0.8\linewidth]{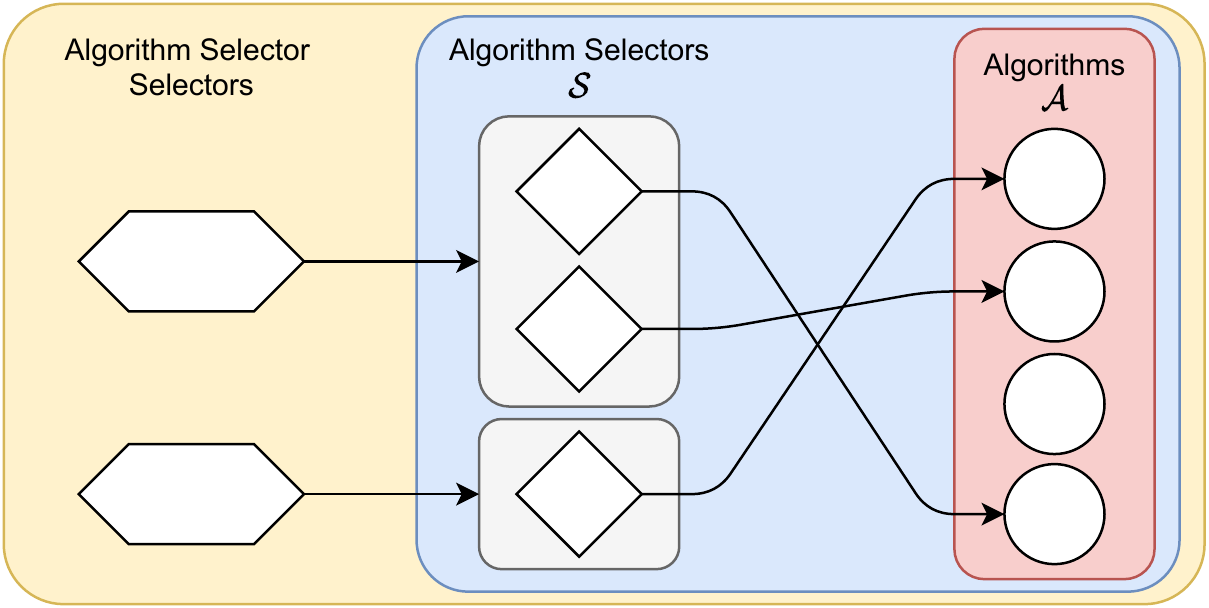}
    \caption{Illustration of the connection between algorithms ($\mathcal{A}$), algorithm selectors ($\mathcal{S}$) and algorithm selector selectors. Algorithms solve instances of an algorithmic problem, whereas algorithm selectors are mappings from an instance to a \textit{single} algorithm from $\mathcal{A}$. Algorithm selector selectors select \textit{one or multiple} algorithm selectors, which in turn each select an algorithm. These selections are then aggregated using an aggregation function (not displayed here).}
    \label{fig:as_vs_meta_as}
\end{figure}

\autoref{fig:as_vs_meta_as} illustrates the relation between algorithms, algorithm selectors and algorithm selector selectors. In the following, we present several instantiations of this framework.

\section{Selecting Single Algorithm Selectors through Meta Learning}\label{sec:meta-learning_ass}

The arguably simplest solution to the meta AS problem is achieved through meta learning \cite{meta_learning_survey}, namely to learn which algorithm selector takes the best decision for a given instance. More formally, one could seek to learn a map
\begin{equation}\label{eq:meta_learn_as}
    s_\mathit{meta}: \mathcal{I} \longrightarrow \selectors \, ,
\end{equation} 
such that the chosen selector returns the most suitable algorithm for a given instance $i$, i.e.,
\begin{equation}
    \left(s_\mathit{meta}(i)\right)(i) \in \arg\min_{a \in \algorithms} \mathbb{E}\left[m(i,a)\right] \,\,\,.
\end{equation} 
In this case, the co-domain of the function $ass$ in (\ref{eq:ass}) is effectively restricted to singleton sets $ass(i) = \{ s \} \in \mathcal{S}$ consisting of only a single algorithm selector $s$\,---\,we shall discuss the consequences of this self-imposed restriction in \autoref{sec:oracle_and_sbs_implications}. Moreover, the aggregation $agg$ in (\ref{eq:agg}) is the identity, or, stated differently, there is actually no need for learning an aggregation function. 

Observe that this approach is essentially a special case of the standard AS problem itself, with a very specific set of algorithms to choose from, namely algorithm selectors. Hence, standard AS methods (see \autoref{subsec:as_solutions}) can in principle be applied. It is important to note that algorithm selection approaches not relying on a feature representation of instances do not necessarily have an advantage in terms of runtime anymore, because they may select an algorithm selector which in turn requires the feature representation. If the feature computation has to be performed either on the meta or on the base level, its time has to be taken into account as well. However, there is no need to perform the computation twice, if both the algorithm selector selector and the algorithm selector require it, because the resulting features can be shared.

\subsection{Limits of learned algorithm selector selection}\label{sec:oracle_and_sbs_implications}
Limiting ourselves to choosing only a single algorithm selector for a given instance instead of leveraging multiple ones obviously has consequences in terms of achievable algorithm selection performance. To elaborate on these consequences, let us define an algorithm selector oracle (AS-oracle) as 
\begin{equation}
    \mathit{ass}^*(i) \in \arg\min_{s \in \selectors} \mathbb{E}\left[m(i,s(i))\right] \,\,\, .
\end{equation} 
It is important to note that the AS-oracle is in general not identical to the oracle on the base level, as the set of algorithms to choose from may change. For a better understanding, consider an example with two algorithms $a_1$ and $a_2$ and two algorithm selectors $s_1$ and $s_2$, where both always select algorithm $a_1$. Furthermore, assume there exists an instance for which $a_2$ performs better than $a_1$, and hence the oracle would select $a_2$. However, the AS-oracle can only select $s_1$ or $s_2$, which in turn both select $a_1$, resulting in a decrease in oracle performance.

Generally speaking, in order to preserve the original oracle, it is necessary that, for each instance, at least one algorithm selector exists that selects the best algorithm for that instance. Otherwise, the AS-oracle performance may degrade compared to the oracle. In practice, there will be at least one such instance most of the time, and hence an important question is how much the oracle performance degrades. As we show in our experimental evaluation, the degradation strongly depends on the scenario at hand, and ranges from less than $1\%$ to over $85\%$. 

Similarly to the oracle, the SBS on the meta level changes as well, since the single best algorithm selector (SBAS), i.e., the algorithm selector which is best on average, is now an algorithm selector, making it a lot stronger baseline than the single best solver. Hence, while the SBS selects the actual problem solving algorithm that is best on average and accordingly does not depend on instance features, the SBAS does in fact depend on such features as long as it is not identical to the SBS. Observe that this results in a significant disadvantage for the SBAS in terms of achievable \textit{PAR10} scores due to the time required to compute these instance features.

Obviously, these implications also influence the performance gains that can be achieved by algorithm selector selectors of the form (\ref{eq:meta_learn_as}) in comparison to algorithm selectors. As the oracle performance most likely degrades, while the SBS performance most likely improves, the gap between the two also decreases, offering less potential for algorithm selection approaches to close this gap.

\section{Constructing Ensembles of Algorithm Selectors}\label{sec:ensembles}

As mentioned earlier, the restriction to choose only a single algorithm selector seems like an unnecessary constraint and may even lead to a potential loss in achievable algorithm selection performance. Accordingly, one may think about using a \textit{composition} of algorithm selectors, which can play to their strengths on some instances while compensating for each other's weaknesses on other instances. This idea motivates us to construct a mapping of the form (\ref{eq:ass}) through \textit{ensemble learning}.

\begin{figure}[t]
    \centering
    \includegraphics[width=0.70\textwidth]{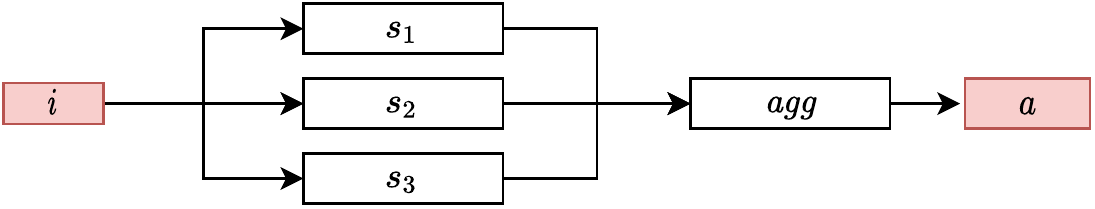}
    \caption{This figure depicts the general process of predicting / selecting an algorithm for a given instance through a trained ensemble of algorithm selectors $s_1, s_2, s_3$.}
    \label{fig:ensemble}
\end{figure}

Ensemble learning \cite{dietterich00_ensembles} presumably constitutes the most natural technique to combine several machine learning approaches into a joint one, with the goal to improve in performance. In algorithm selection, an ensemble can be thought of as a set of algorithm selectors $\selectors$, called \textit{base algorithm selectors}, which are either trained independently or dependently on each other. At prediction time, each selector is queried for the given instance $i$, and the algorithm choices are aggregated into a final choice using an aggregation function as defined in (\ref{eq:agg}). The concrete strategy used to make the selectors cooperate depends on the ensemble technique being used. \autoref{fig:ensemble} depicts the general process of predicting / selecting an algorithm for a given instance through a trained ensemble of algorithm selectors.

As mentioned earlier, allowing for the selection of multiple algorithm selectors also requires the definition of an aggregation function in order to finally return a single algorithm. In principle, the aggregation functions can either depend on the instance, i.e., are instance-specific, or can be fixed across instances. Similarly, they can either be learned or be predefined. 

In general, to be successful, ensembles require a certain degree of heterogeneity of the predictions. Therefore, the different algorithm selectors should not always coincide in their selections. Otherwise, it can easily happen that the majority of predictions made by the base selectors are identical. Hence, in such a situation, the prevalent selector (maybe with slight but negligible variations) dominates the predictions of the entire ensemble, only yielding a computationally more expensive variant of the respective dominating selector.
To avoid this problem, most ensemble methods strive for a heterogeneous set of base selectors. This can be achieved through a suitable choice of base selectors given to the method, like for example in voting. Alternatively, in the case of methods such as bagging, which only work with a single base selector, different variants of the same selector can be trained on different data sets.

Intuitively, the training and querying of more than one selector might be counter-intuitive in settings where runtime is optimized, as it automatically results in larger runtime. In this regard, it is important to note that the majority of the runtime is required for training the selectors in the ensembles. In AS, we can assume this training to be performed \textit{offline}, i.e., prior to the actual selection of algorithms. Hence, longer training times do not constitute a real disadvantage, as long as prediction (querying the ensemble members) remains fast.

In the following, we first elaborate on different aggregation strategies. Although some of these aggregation functions include learnable components, they are fixed across instances, i.e., the aggregation of predictions does not depend on the query instance. Then, we present several ensemble techniques for creating a pool of algorithm selectors, in particular voting \cite{dietterich00_ensembles}, bagging \cite{breiman96b_bagging}, and boosting \cite{schapire1990strength_boosting}. We continue with a discussion of stacking \cite{wolpert92_stacking}, which can be seen as a \textit{learned, instance-specific} aggregation method. As such, it is somehow positioned in-between ensemble and meta learning. Finally, we close this section with a methodological comparison of the presented approaches.

\subsection{Aggregation Strategies}
One of the most natural forms of aggregation in our context is \textit{(weighted) majority aggregation}. As the name suggests, it aggregates the algorithm choices by selecting the algorithm that was selected most frequently, potentially weighting the choices of the selectors differently. This is motivated by the idea that selectors with a strong performance should potentially be trusted more than weaker ones. More formally, weighted majority aggregation can be defined as\footnote{$\llbracket \cdot \rrbracket$ denotes the indicator function evaluating to $1$ if the expression is true, and to $0$ otherwise}
\begin{equation}\label{eq:weighted_majority_voting}
    \mathit{agg}_\mathit{(w)maj}(i, \selectors) = \arg\max_{a \in \algorithms} \sum\limits_{s \in \selectors} w_s \cdot \llbracket s(i) = a \rrbracket \, ,
\end{equation} 
where $w_s \in \mathbb{R}^+$ denotes the weight associated with selector $s$. With $w_s = 1$ for all $s \in \selectors$, we recover standard majority voting. To obtain proper weights, a plethora of methods are applicable in principle. However, we simply consider the \textit{nPAR10} score of the different base algorithm selectors on the training data in order to determine corresponding weights\,---\,conducting a cross-validation on the training data for the same purpose turned out to result in similar performance while being computationally more expensive.

Up to now, we assumed that an algorithm selector only returns a single algorithm. While this is typically true in practice, the majority of approaches internally feature more nuanced predictions, often constituting some kind of loss (or score) for each algorithm in $\algorithms$. Accordingly, instead of using only a concrete algorithm choice as the output of the algorithm selectors, we adapted them to return such nuanced predictions where possible. 

More formally, let us assume that each \textit{trained} algorithm selector $s \in \selectors$ cannot only be evaluated on $i \in \mathcal{I}$, but that it also allows access to $\widehat{m}_s(i,a)$, i.e., to the corresponding internal score of each algorithm $a \in \algorithms$. For those approaches where such a score cannot be extracted explicitly, e.g., multi-class algorithm selectors, we define dummy losses as 
\begin{equation}\label{eq:dummy_scores}
    \widehat{m}_s(i,a) = 
    \begin{cases}
        0 & \text{if $s(i) = a$} \\
        1 & \text{else}
    \end{cases} 
\end{equation} 
for all instances $i \in \mathcal{I}$ and algorithms $a \in \mathcal{A}$, such that all approaches can be assumed to work as defined in (\ref{eq:canonical_as}). 

With this consideration, aggregations on this more nuanced level of scores instead of the level of final choices can be made. The most straight-forward aggregation function on this level is the \textit{arithmetic mean}, i.e.,
\begin{equation}
    \mathit{agg}_\mathit{avg}(i, \selectors) = \arg\min_{a \in \algorithms} \frac{1}{\vert \selectors \vert}\sum\limits_{s \in \selectors} \widehat{m}_s(i,a) \,.
\end{equation} 
While conceptually simple, it requires the performance surrogates of the different selectors to approximate the same function. Otherwise, the predictions are incomparable, and averaging is not a meaningful operation. For example, combining the output of a ranking loss function optimized by one selector with the estimated average \textit{PAR10} scores of another does not make any sense. In principle, the arithmetic mean can also be turned into a weighted version as done in (\ref{eq:weighted_majority_voting}).

In order to be able to aggregate on this more nuanced level while overcoming the weakness of the arithmetic mean, we propose to aggregate \textit{rankings} (rank aggregation) of algorithms constructed from the algorithm scores obtained from the selectors. More precisely, we can assume that each selector $s$ returns a ranking over the algorithms in $\algorithms$ by sorting them in increasing order w.r.t.\ $\widehat{m}_s(i, \cdot)$, such that the presumably best algorithm is put on the first position in the ranking, the second-best on the second position, etc. Having obtained such a ranking over the algorithms for each selector, they need to be aggregated in order to draw a conclusion and eventually return a single algorithm as the final choice.

A very simple method for rank aggregation is called \textit{Borda count} \cite{borda1784memoire_borda_count}. Given a ranking of $n$ items, it assigns $n$ points to the top item, $n-1$ points to the second-best, and so forth. This is done for each ranking to be aggregated, and the consensus ranking is obtained by sorting the items (algorithms in our case) in descending order according to their total sum of points. 
As pointed out by \cite{dworkKNS01_rank_aggregation}, the Borda count has a number of less appealing properties, at least from a theoretical point of view. On the other side, its linear time complexity makes it fast to compute. This is in sharp contrast to other rank aggregation techniques that involve intractable optimization problems \cite{dworkKNS01_rank_aggregation}. Besides, Borda comes with provable approximation guarantees for several other aggregation techniques \cite{copp_ob06}. Overall, it seems to be a good compromise for the case of algorithm selection, where predictions are performed under tight time constraints. 

Formally, we can use Borda count as an aggregation function for our setting as follows, where $\mathit{rank}: \mathcal{I} \times \selectors \times \algorithms \to \mathbb{N}$ returns the rank of an algorithm $a$ in the ranking returned by a selector $s$ on an instance $i$:
\begin{equation}
    \mathit{agg}_\mathit{borda}(i, \selectors) = \arg\min_{a \in \algorithms} \sum\limits_{s \in \selectors} \mathit{rank}(i, s, a)
\end{equation} 
Ties are handled by assigning to all tied algorithms the average of the block of ranks they occupy \cite{saari2000mathematics_borda_extension}. In practice, ties can only be caused through the dummy scores introduced in (\ref{eq:dummy_scores}). Therefore, they always occur at the end of the rankings. Theoretically, identical scores of $\widehat{m}(i, \cdot)$ could also result in ties, but this never happened in practice. 

While the aggregation techniques outlined above appear to be meaningful in the context of the algorithm selection task, we would like to point out that other aggregation techniques are of course conceivable and could be used instead.

\subsection{Voting}

\begin{figure}[t]
    \centering
    \includegraphics[width=0.40\textwidth]{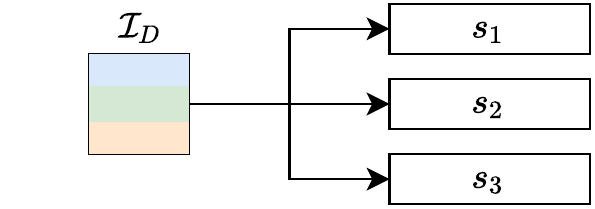}
    \caption{This figure depicts the training process of a voting ensemble, where each base algorithm selector is trained with the same training instances. Ensemble heterogeneity is achieved by choosing a heterogeneous set of algorithm selectors in advance.}
    \label{fig:voting}
\end{figure}

Voting ensembles are presumably the easiest form of ensemble learning: Each algorithm selector in a set $\selectors' \subseteq \selectors$ is trained independently of the others on the same training data $\mathcal{I}_D$. At prediction time, all algorithm selectors in $\selectors'$ are queried, and the predictions are aggregated using one of the previously described aggregation strategies. \autoref{fig:voting} depicts the training process of a voting ensemble.

As we demonstrate empirically, it is important to optimize the ensemble composition, i.e., the set of base algorithm selectors $\selectors' \subseteq \selectors$ specifying the ensemble, because the performance of a voting ensemble solely depends on this configurable parameter. Intuitively, a complete evaluation of each possible composition to check the corresponding performance might seem intractable due to the exponential (in $| \selectors |$) number of compositions. However, all base algorithm selectors can be trained on the training data once, so that, in order to estimate the performance of an ensemble composition, only the predictions of the used selectors need to be obtained and aggregated. As the training of the selectors has to be performed only once at the beginning, and the computation of both the predictions and the aggregation can be performed in a negligible amount of time, the evaluation of all possible compositions is feasible as long as the set of algorithm selectors remains moderately large. For example, computing the training performance of each possible voting ensemble composed of up to 7 algorithm selectors required less than 10 hours for all scenarios presented in \autoref{sec:evaluation}. If the size of algorithm selectors becomes larger, more sophisticated optimization methods such as genetic algorithms can be used to find good compositions.

\subsection{Bagging}

\begin{figure}[t]
    \centering
    \includegraphics[width=0.45\textwidth]{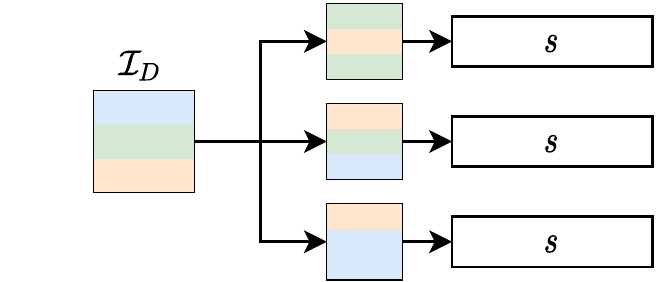}
    \caption{This figure depicts the training process of a bagging ensemble consisting of several instantiations of the same base algorithm selector trained on bootstrapped versions of the original training data.}
    \label{fig:bagging}
\end{figure}

In contrast to voting, \textit{bagging}\footnote{The term is short for short for ``bootstrap aggregating''.} \cite{breiman96b_bagging} only leverages a single kind of algorithm (selector). Therefore, heterogeneity between the ensemble members has to be achieved through data manipulation techniques. To this end, bagging leverages a data resampling technique from statistics called \textit{bootstrapping}, which works as follows. Given a set of training instances $\mathcal{I}_D$ of size $N = \vert \mathcal{I}_D \vert$, it creates a new training instance set by sampling $N$ times from $\mathcal{I}_D$ \textit{with replacement}. The actual ensemble is constructed by sampling $k$ such new training instance sets $\mathcal{I}_D^{(1)}, \ldots, \mathcal{I}_D^{(k)}$ and training one instantiation of the provided algorithm selector on each of the $k$ training sets. Thus, the ensemble eventually consists of $k$ algorithm selector instances. At prediction time, one of the previously discussed aggregation functions can be used to aggregate the predictions (selections) of the different selectors. \autoref{fig:bagging} depicts the training process of a bagging ensemble.

We would like to point out that we bootstrap on the level of the problem instances and not on the level of the actual training data points ((instance/algorithm)-pairs or (instance/algorithm performance)-pairs). This is done in order to allow the selection algorithms themselves to construct their training data points. In principle, this may lead to differently large training data sets for the corresponding base algorithm selectors if the number of training performance values $m(i,\cdot)$ varies across instances. However, we assume that either $m(i,a)$ is available or we know at least that $m(i,a) > C$ for all $i \in \mathcal{I}_D, a \in \algorithms$, and hence can reasonably impute these missing values, thereby solving the problem of differently sized training data sets.

\subsection{Boosting}
\begin{figure}[t]
    \centering
    \includegraphics[width=0.70\textwidth]{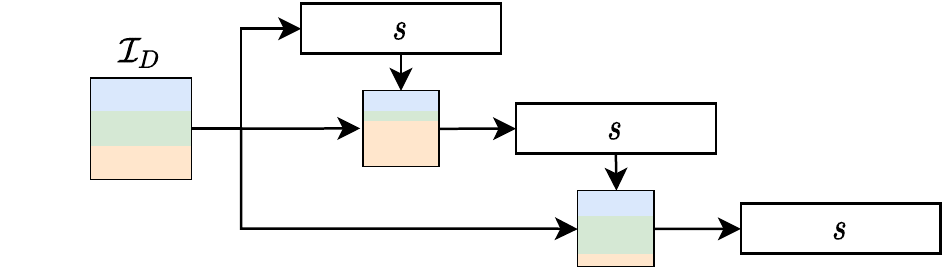}
    \caption{This figure depicts the training process of a boosting ensemble. Similar to bagging, the ensemble constitutes several instances of the same base algorithm selector. These are subsequently trained  on differently weighted versions of the training data.}
    \label{fig:boosting}
\end{figure}

While both voting and bagging fit ensemble members independently of each other (except for (partially) identical training data), boosting \textit{successively} trains its members, each time re-weighting the training instances \cite{schapire1990strength_boosting}. After each iteration, i.e., trained selector, the error of the previous selectors is determined and more weight is put onto those instances where a wrong algorithm selection has been performed, while the weight on correctly judged instances is reduced. Similar to bagging, boosting only uses a single selector as a basis of which it trains instantiations based on differently weighted versions of the same training instance set in order to achieve diversity w.r.t.\ its ensemble members. At prediction time, the predictions of each of the trained selectors are obtained and combined into a joint prediction using a weighted aggregation, using the weights that have been determined as part of the boosting algorithm during the training phase. \autoref{fig:boosting} illustrates the training process of a boosting ensemble.

In boosting algorithms for multi-class classification, such as SAMME \cite{hastie2009multi}, and regression problems, such as AdaBoost.R2 \cite{drucker1997improving}, one would naturally consider multi-class classification errors and regression losses, respectively, for re-weighting training instances. However, due to the inferior performance of AdaBoost.R2 in preliminary experiments, we focus on SAMME for the remainder of this paper.

\subsection{Stacking}
\begin{figure}[t]
    \centering
    \includegraphics[width=\textwidth]{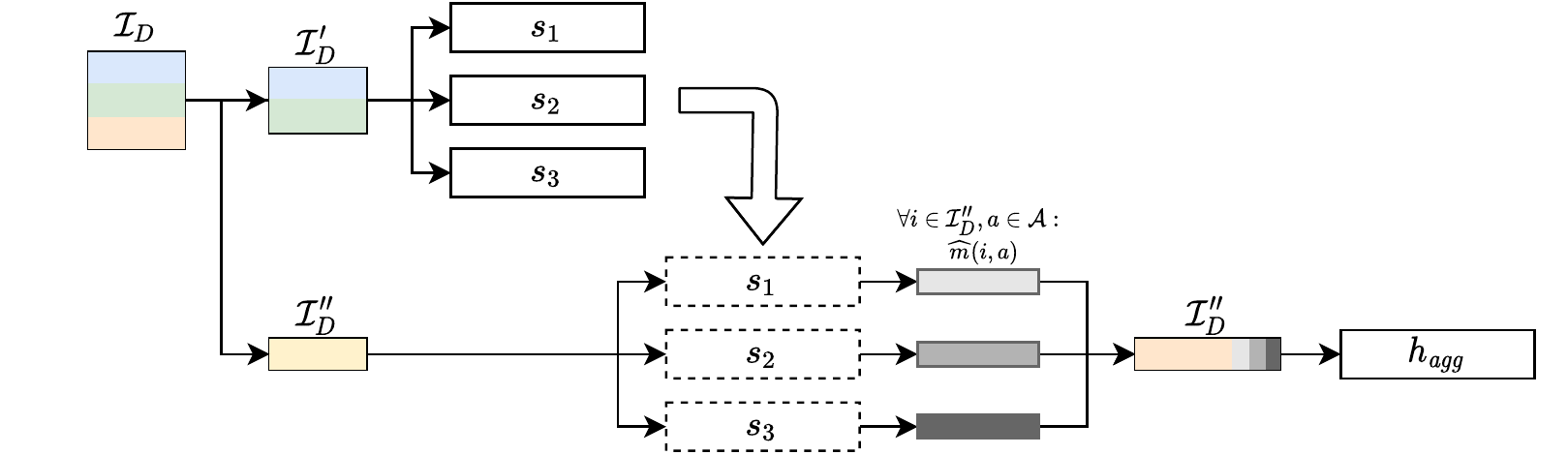}
    \caption{This figure depicts the general idea behind a stacking ensemble. Each ensemble member is trained with the same subset of training instances and the remaining instances are augmented with the corresponding predictions of the trained selectors. Then, a meta-learner, i.e. an additional algorithm selector, $h_\mathit{agg}$ is trained on this augmented data, which decides on the algorithm to select.}
    \label{fig:stacking}
\end{figure}

In the previous ensemble techniques, the aggregation strategy is always fixed from the beginning and independent of the actual instance at hand. The idea of stacking is to learn the aggregation, i.e., how to best aggregate the predictions of the base algorithm selectors for a given instance. Therefore, a meta-learner 
\begin{equation}
    \mathbf{h}_{\mathit{agg}}: \mathcal{I} \times \mathbb{R}^{\vert\selectors\vert \times \vert\algorithms\vert} \rightarrow \algorithms
\end{equation}
is fitted and used to aggregate the predicted performances $\widehat{m}(i,a)$ of each algorithm selector $s \in \selectors$ for a given instance $i \in \mathcal{I}$ and each algorithm $a \in \algorithms$ into a joint decision. To avoid any bias in the training data for the meta-learner, it needs to be ensured that this data is disjoint from the training data of the base algorithm selectors. Therefore, the set of training instances $\mathcal{I}_D$ is normally split into a set of base algorithm selector training instances $\mathcal{I}'_D \subset \mathcal{I}_D$ and a set of meta-learner training instances $\mathcal{I}''_D \subset \mathcal{I}_D$ such that $\mathcal{I}'_D \cap \mathcal{I}''_D = \emptyset$.\footnote{Although theoretically correct, we did actually not do that split in our experimental evaluation in Section \ref{sec:evaluation}, where is led to worse empirical performance.} As all possible base algorithm selectors are used, each can be trained independently on the same subset of training instances $\mathcal{I}'_D$ as a first step such that the training data for the meta-learner can be built. Then, the meta-learner is trained based on the features $f(i) \in \mathbb{R}^d$ of each training instance $i \in \mathcal{I}''_D$ extended by the predictions $\widehat{m}_s(i,\cdot)$ of all base algorithm selectors $s \in \selectors$ on these instances. 
At prediction time, each base algorithm selector $s \in \selectors$ is queried, its predictions $\widehat{m}_s(i,\cdot)$ are concatenated and attached to the instance features $f(i) \in \mathbb{R}^d$ of instance $i$, based on which the meta-learner predicts which algorithm to choose. As the meta-learner is an algorithm selector itself, any of the base algorithm selectors can be used. \autoref{fig:stacking} depicts the general idea of a stacking ensemble.

Since stacking is working on an (extended) feature representation, standard feature selection techniques can be used to reduce the number of features and help the meta-learner achieve better prediction performance. Thus, the ensemble composition does not require any optimization upfront. For an overview of feature selection methods, we refer to \cite{guyonE03_feature_selection}.

\subsection{Comparison of the Approaches}
\begin{figure}[t]
    \centering
    \includegraphics[width=\textwidth]{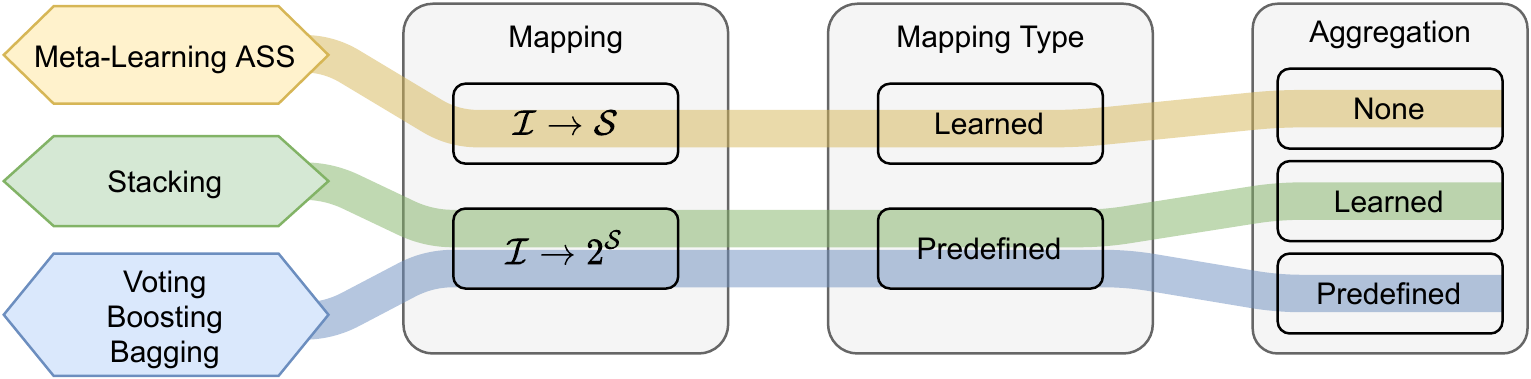}
    \caption{Illustration of the different approaches w.r.t.\ the kind of mapping they model, how this mapping is constructed, and how the required aggregation is obtained.}
    \label{fig:approach_relation}
\end{figure}

To put the approaches presented so far into the broader context of meta AS, we close this section by revisiting them w.r.t.\ to their most important properties. \autoref{fig:approach_relation} provides an overview and illustrates how the approaches relate to each other. It clarifies what kind of mapping these approaches model, how this mapping is constructed, and how the required aggregation function is constructed.

As an important observation, note that some approaches involve learning on the meta level while others do not. The former most obviously holds for learning an algorithm selector selector (cf. \autoref{sec:meta-learning_ass}), where the modeled mapping is learned directly. On the other side, most ensemble approaches (cf. \autoref{sec:ensembles}) do not require any learning on the meta level, because their mapping is essentially predefined. Stacking is somehow in-between these two groups: the mapping itself is predefined, but the aggregation function is learned on the meta level.

\section{Experimental Evaluation}\label{sec:evaluation}

In this section, we provide an empirical evaluation of the ideas presented in the preceding sections. It is organized into four main parts. First, we introduce our experiment setup. Second, we investigate the chance for performance improvements when learning algorithm selector selectors and evaluate the performance of standard algorithm selectors working as algorithm selector selectors. Third, we evaluate the performance of the different ensemble methods presented earlier and discuss the results. We end this section by drawing a broader conclusion from these results.

\subsection{Experiment Setup}
All evaluations are run on a subset of the scenarios from the ASlib v4.0 benchmark suite \cite{bischlKKLMFHHLT16} with a 10-fold cross-validation, where the folds are provided by the benchmark. Table \ref{tab:scenarios} shows the scenarios used with their corresponding characteristics.
\begin{table}[ht]
	\centering
	\caption{Overview of examined ASlib scenarios including their number of instances (\#I), unsolved instances (\#U), algorithms (\#A), provided features (\#F), and the cutoffs (C).}

\includestandalone[width=\textwidth]{figures/scenarios}

  \label{tab:scenarios}
\end{table}

The performance of the approaches is measured in terms of the \textit{normalized penalized average runtime} (\textit{nPAR10}) metric as defined in (\ref{eq:npar10}) if not mentioned otherwise. Recall that a value of $0$ indicates oracle performance, values below $1$ an improvement over the SBS, and values above $1$ a degradation compared to the SBS. To allow for a better visual interpretation, we sometimes illustrate results aggregated over all scenarios. Needless to say, such aggregations have to be treated with care, because (differences between) performance degrees are not easily comparable across scenarios.

The set of algorithm selectors used for the evaluation consists of $\selectors = \{$PerAlgo, SATzilla'11, R2S-Exp, R2S-PAR10, SUNNY, ISAC, Multiclass$\}$, which all have been described in \autoref{sec:algorithm-selection}. These are used both as meta learners, but also as base algorithm selectors for the ensembles. Furthermore, we compare all ensemble variants against the \textit{single best algorithm selector} (SBAS), i.e., the algorithm selector which performs best across all scenarios in terms of average or median \textit{nPAR10} performance.

All experiments were run on machines featuring Intel Xeon E5-2695v4@2.1GHz CPUs with 16 cores and 64GB RAM.
In the interest of reproducibility of our results, all code, including detailed documentation of the experiments and execution instructions, is available at GitHub\footnote{ \url{https://github.com/alexandertornede/as_on_a_meta_level}}.

\subsection{Meta Learning for Selecting an Algorithm Selector}

\begin{figure}[t]
    \centering
    \includegraphics[width=0.9\linewidth]{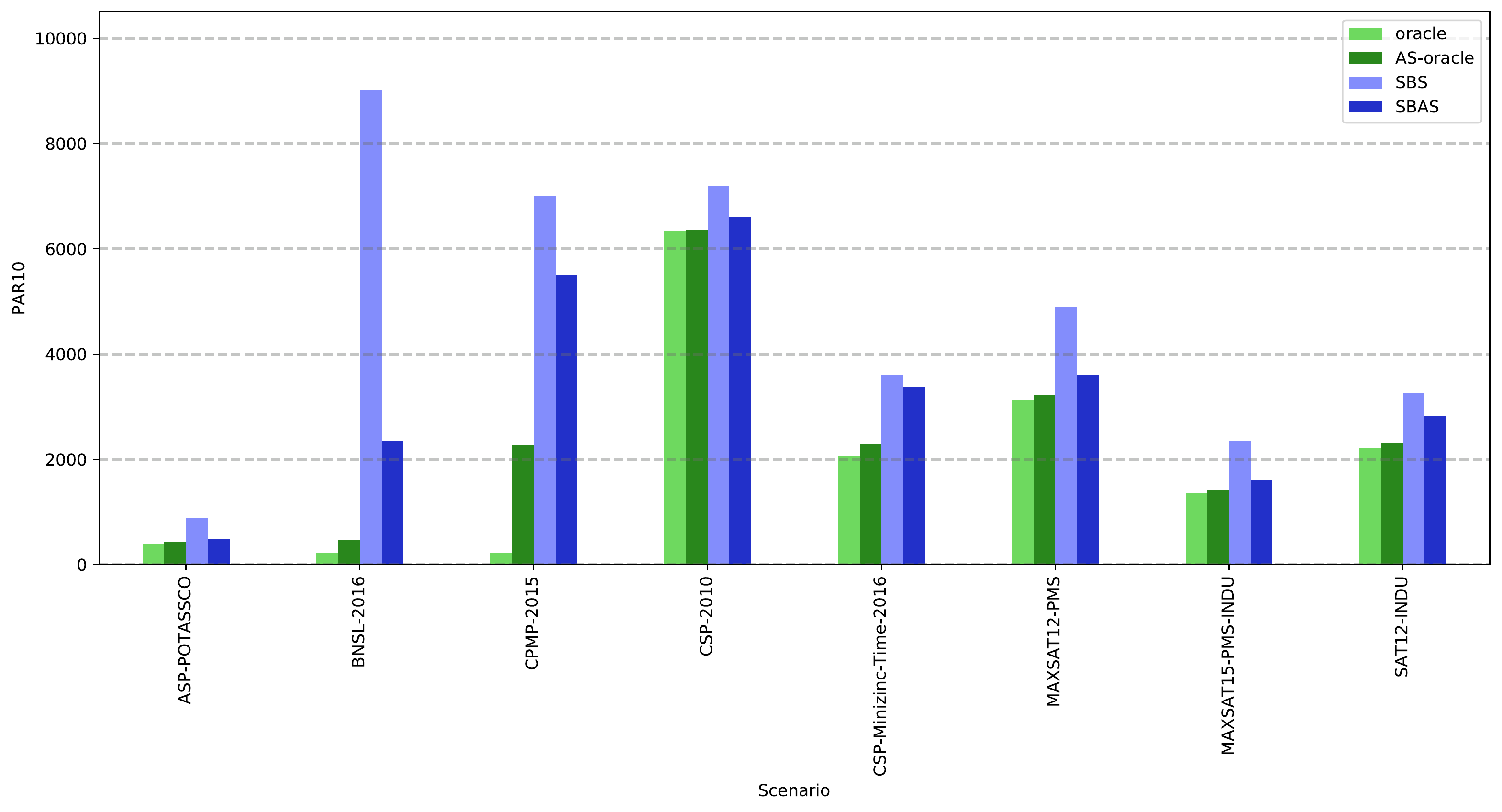}
    \caption{This figures shows the PAR10 scores of the oracle, AS-oracle, SBS and SBAS on a subset of the ASlib v4.0 benchmark scenarios as bar charts.}
    \label{fig:sbs_vbs_comparison}
\end{figure}

\autoref{fig:sbs_vbs_comparison} shows the PAR10 scores of the oracle, AS-oracle, SBS and SBAS on a subset of the ASlib v4.0 benchmark scenarios. As one can see, several of the implications we noted in \autoref{sec:oracle_and_sbs_implications} can be validated empirically. Firstly and most importantly, although the SBS/oracle gap is a lot larger than the SBAS/AS-oracle gap, the SBAS/AS-oracle gaps are non-negligible, and hence constructing an algorithm selector selector can in principle make sense. For example, consider scenarios BNSL-2016 or CPMP-2015 with large SBAS/AS-oracle gaps.

As we noted earlier, the reason why these gaps become smaller is that the oracle performance degrades when moving to the meta level for all scenarios, whereas the SBS performance tends to improve, because the SBAS is essentially an algorithm selector. While the degradation in oracle performance is moderate for the majority of scenarios (less than $10\%$), the improvement of the SBAS over the SBS is non-negligible, as the more successful the algorithm selectors considered by the algorithm selector selectors are, the larger this performance gain is.

\begin{table}[t]
    \centering
    \caption{PAR10 scores of all base- and algorithm selector selectors normalized wrt. the standard oracle and SBS. The result of the best approach is marked in bold for each scenario. Moreover, for the meta-algorithm selectors the values in brackets $(a/b)$ indicate that that the approach achieves a performance better or equal to $a$ base-approaches and is worse than $b$ base-approaches.}
    \label{tbl:n_par10_level0}
    \resizebox{\textwidth}{!}{
        \input{tables/n_par_10_all_normalized_by_level_0.tex}
    }
\end{table}

Table \ref{tbl:n_par10_level0} shows the \textit{nPAR10} scores of all algorithm selectors and the corresponding algorithm selector selectors of form (\ref{eq:meta_learn_as}). Moreover, for the algorithm selector selectors, the values in brackets $(a/b)$ indicate that the approach achieves a performance better or equal to $a$ base approaches and is worse than $b$ base approaches.

Unsurprisingly, most algorithm selector selectors are able to consistently improve over the SBS. However, moving to the meta level proves to be beneficial for only four scenarios and these improvements are even distributed across different algorithm selector selectors. To explain this moderate result, we speculate that the considered AS approaches are not able to unleash their full potential on the meta level, although considerable SBAS/AS-oracle gaps exist, as we have seen previously. However, the win/loss scores in brackets indicate that moving to the meta level is beneficial in the sense that a more robust performance across several scenarios can be achieved.



\subsection{Voting Ensembles}
\begin{figure}[t]
    \centering
    \includegraphics[width=\textwidth]{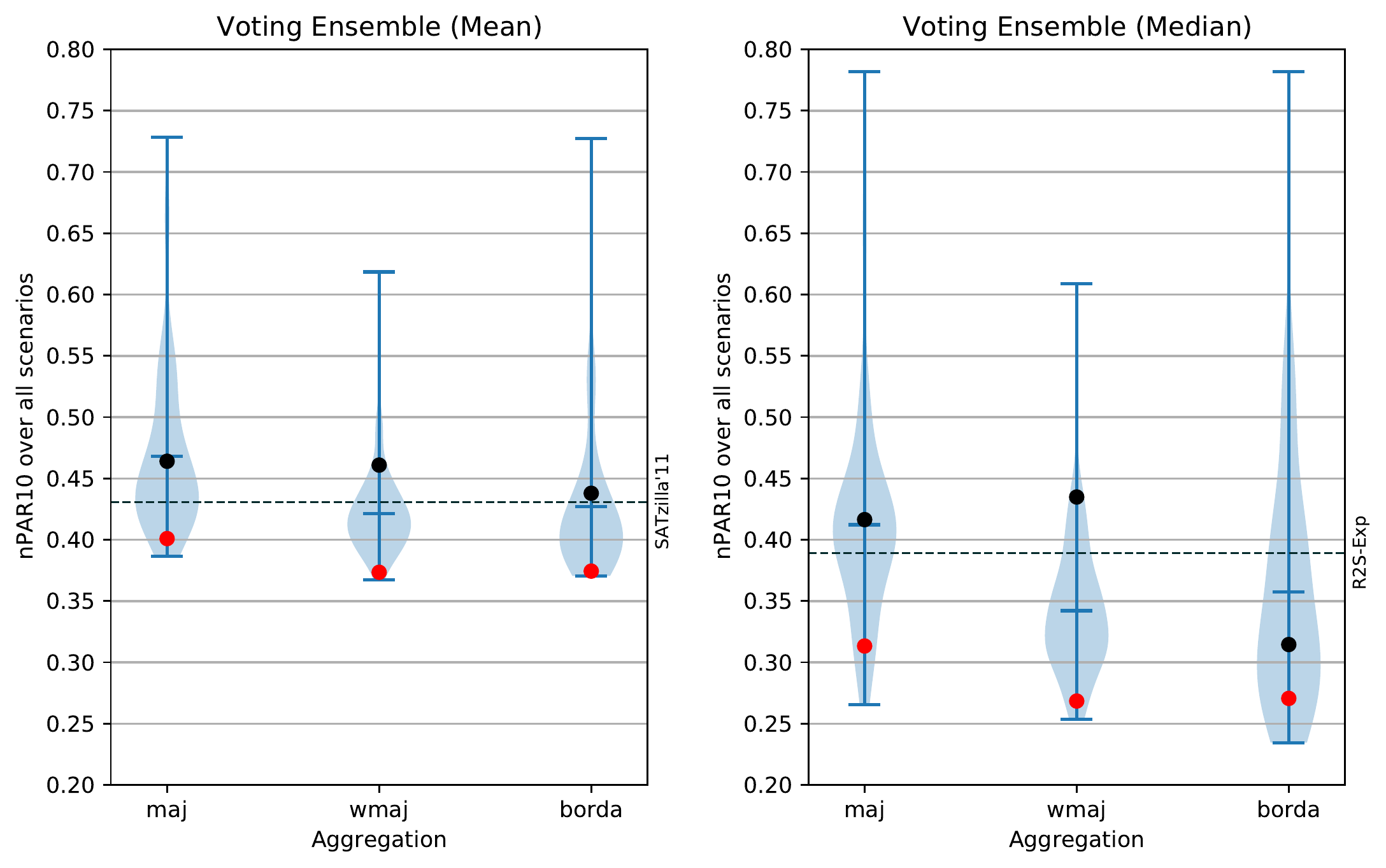}
    \caption{Mean / median performance in terms of \textit{nPAR10} (over all scenarios) of all possible voting ensemble compositions as violin plots grouped by the aggregation strategy being used. The dashed line indicates the performance of the SBAS, the black dot indicates the performance of the best composition w.r.t.\ to the training performance, whereas the red dot indicates the performance of the ensemble with all base algorithm selectors.}
    \label{fig:voting_violin_plot}
\end{figure}

\autoref{fig:voting_violin_plot} shows the average / median performance in terms of \textit{nPAR10} (over all scenarios) of all possible voting ensemble compositions as violin plots grouped by the aggregation strategy being used. The dashed line indicates the performance of SBAS, the black dot indicates the performance of the best composition w.r.t.~the training performance, whereas the red dot indicates the performance of the ensemble with all base algorithm selectors.

First of all, it is important to note that voting ensembles offer a lot of optimization potential in terms of both mean and median performance in comparison to the SBAS. While a concrete optimization of the ensemble composition (black dots) does not seem to be beneficial, simply using all possible base algorithm selectors as ensemble members often comes close to the lower performance bound of the voting ensemble strategy. Independent of the aggregation strategy, a voting ensemble with all base algorithm selectors is always able to improve over the best single algorithm selector, sometimes even drastically (e.g., Borda aggregation in terms of median performance). Overall, the weighted majority and the Borda aggregation seem to be on a par in terms of performance, while both seem to be superior to the simple majority aggregation.

It is important to understand the scope of the improvement depicted here. Although both SATzilla'11 and R2S-Exp already offer a remarkable performance and represent the state of the art in algorithm selection, they are beaten by around $15\%$ (mean) and $45\%$ (median), which constitute remarkable improvements.

\subsection{Bagging Ensembles}

\begin{figure}[t]
    \centering
    \includegraphics[width=\textwidth]{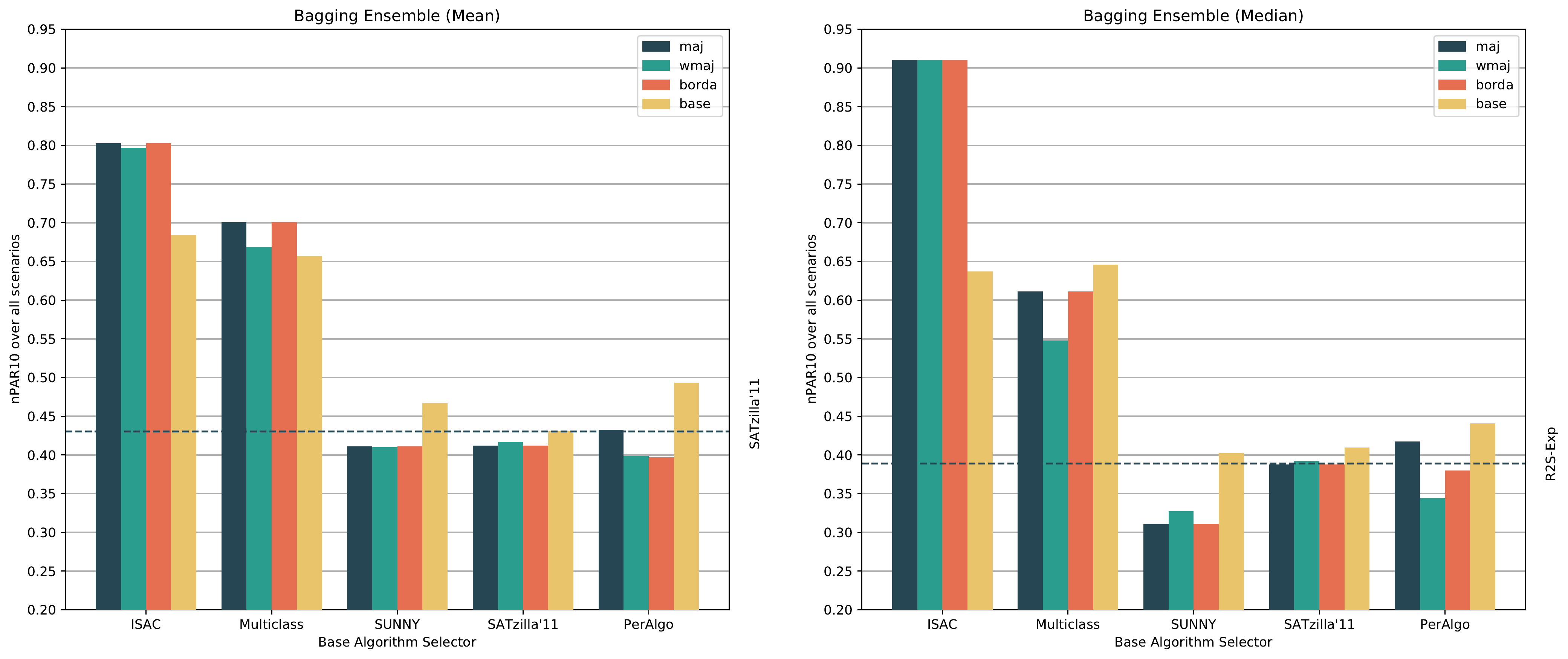}
    \caption{Average / median \textit{nPAR10} performance over all scenarios of each bagging ensemble with 10 instantiations of the corresponding base algorithm selector and different aggregation functions. Moreover, the performance of the corresponding base algorithm selector is shown. Once again, the dashed line indicates the performance of the SBAS.}
    \label{fig:bagging_bar_chart}
\end{figure}

\autoref{fig:bagging_bar_chart} shows the average / median \textit{nPAR10} performance over all scenarios of each bagging ensemble with 10 instantiations of the corresponding base algorithm selector and different aggregation functions. Moreover, the performance of the corresponding base algorithm selector is shown. Once again, the dashed line indicates the performance of the SBAS.

While both ensemble variants equipped with ISAC or Multiclass as a base algorithm selector deteriorate in terms of performance compared to the SBAS, SUNNY, SATzilla'11, and PerAlgo are able to improve both in terms of mean and median performance if the right aggregation is chosen. Surprisingly, none of the aggregation functions seems to be dominating the others. Furthermore, it can be seen that bagging improves the performance of SUNNY, SATzilla'11 and PerAlgo, but mostly worsens the performance for ISAC and offers mixed results for Multiclass. 

In light of the general experience with bagging in machine learning, the performance deterioration of the ISAC ensemble in comparison to its base selector may appear surprising. We conjecture that the negative effect of ensembling is due to the specific characteristics of this method. ISAC applies a clustering technique in order to form clusters over the training instances and computes a threshold $t$ based on the average distances of all instances to their corresponding cluster centroid and the standard deviation over these values. At prediction time, ISAC finds the centroid which is closest to the new instance and returns the algorithm performing best on the cluster, if the distance to the centroid is below the aforementioned threshold. If this is not the case, the SBS is returned. Thus, the threshold can be seen as a fail-safe in case ISAC considers the closest cluster to be too different to draw any reasonable conclusions. After careful investigation, we found that the threshold $t$ decreases for the ensemble members trained on bootstrapped training instance sets as both the average distance and the standard deviation decreases. As a result, the ensemble members mostly deteriorate to the SBS and suggest the SBS on a majority of the instances. This explains the decrease in performance and the similar results of the different aggregation strategies. 

We note that Run2Survive was left out as a base algorithm selector for bagging as it cannot easily be trained with bootstrapped instance training sets on scenarios with many censored samples. In such cases, bootstrapping often leads to training data sets consisting of censored samples only, which the approach cannot handle.

\subsection{Boosting Ensembles}
\begin{figure}[t]
    \centering
    \includegraphics[width=\textwidth]{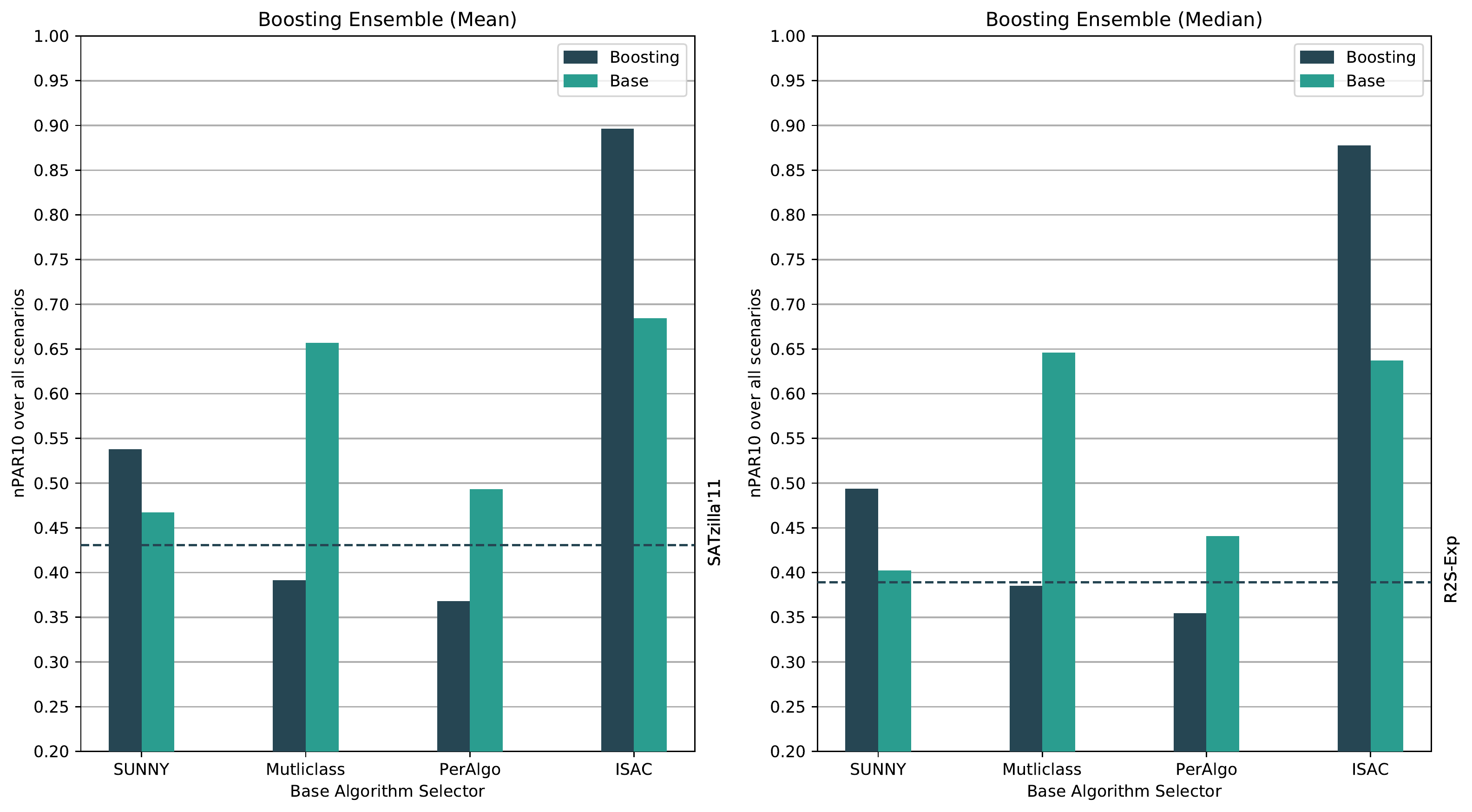}
    \caption{Average / median \textit{nPAR10} performance over all scenarios of each boosting ensemble with 20 iterations and different aggregation functions. Moreover, the performance of the corresponding base algorithm selector is shown. Once again, the dashed line indicates the performance of the SBAS.}
    \label{fig:boosting_bar_chart}
\end{figure}

\autoref{fig:boosting_bar_chart} shows the average / median \textit{nPAR10} performance over all scenarios of each boosting ensemble with 20 iterations and different aggregation functions.

\begin{figure}[t]
    \centering
    \includegraphics[width=\textwidth]{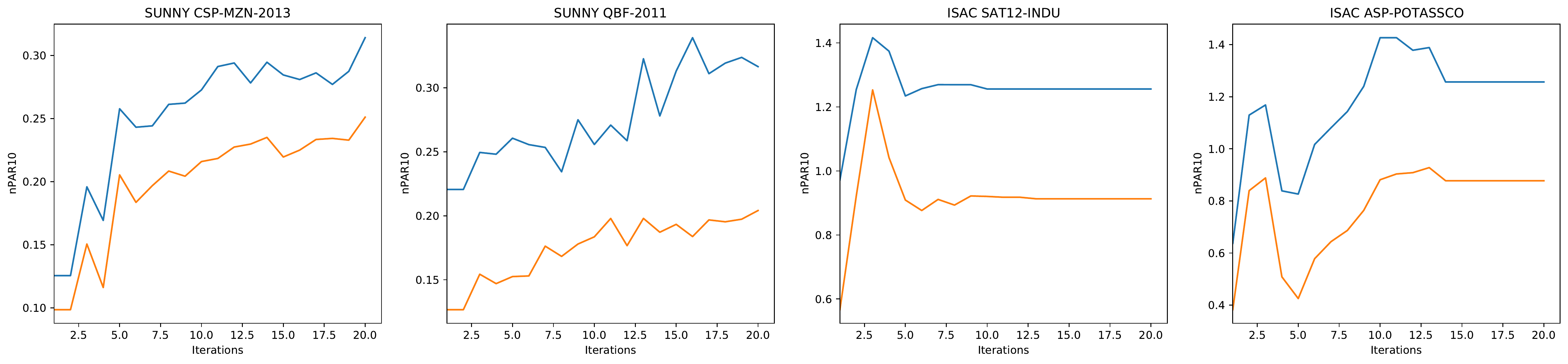}
    \caption{Learning curves featuring training (orange) and testing (blue) \textit{nPAR10} scores of the SAMME boosting algorithm with SUNNY (left two) and ISAC (right two) as a base selector on two instances.}
    \label{fig:boosting_degenerate_learning_curve}
\end{figure}

While the performance of the PerAlgo and Multiclass algorithm selector improve through boosting, the performance of SUNNY and ISAC degrades. Once again, the degradation of ISAC can be explained by the same phenomenon as in the case of bagging: the instance weighting required by boosting was implemented through data sampling, whence ISAC mostly degenerates to the SBS. We chose to do so, since not all of the base algorithm selectors inherently support instance weights, but we wanted to investigate boosting variants powered by as many base algorithm selectors as possible. The degradation of the performance of SUNNY can also be explained in a similar fashion. Recall that SUNNY essentially is a similar $k$-nearest neighbor algorithm, which, given a new instance, returns the algorithm which performs best in terms of PAR10 performance on the $k$ nearest instances in the training data. However, this training data mostly consists of instances with a high weight as all others have a lower chance of being sampled. As a consequence, SUNNY will return the algorithm performing best on average on exactly these instances, while completely ignoring all other instances. This results in degenerate boosting learning curves as depicted in \autoref{fig:boosting_degenerate_learning_curve}. The problem is less dominant for selectors that generalize in a more sophisticated way across the features, such as PerAlgo or Multiclass. For instance-based approaches such as SUNNY or ISAC, different forms of boosting specialized for k-NN approaches \cite{garcia2009boosting} or clustering \cite{frossyniotis2004clustering} might be more promising and should be investigated in future work.

\subsection{Stacking}
\begin{figure}[t]
    \centering
    \includegraphics[width=\textwidth]{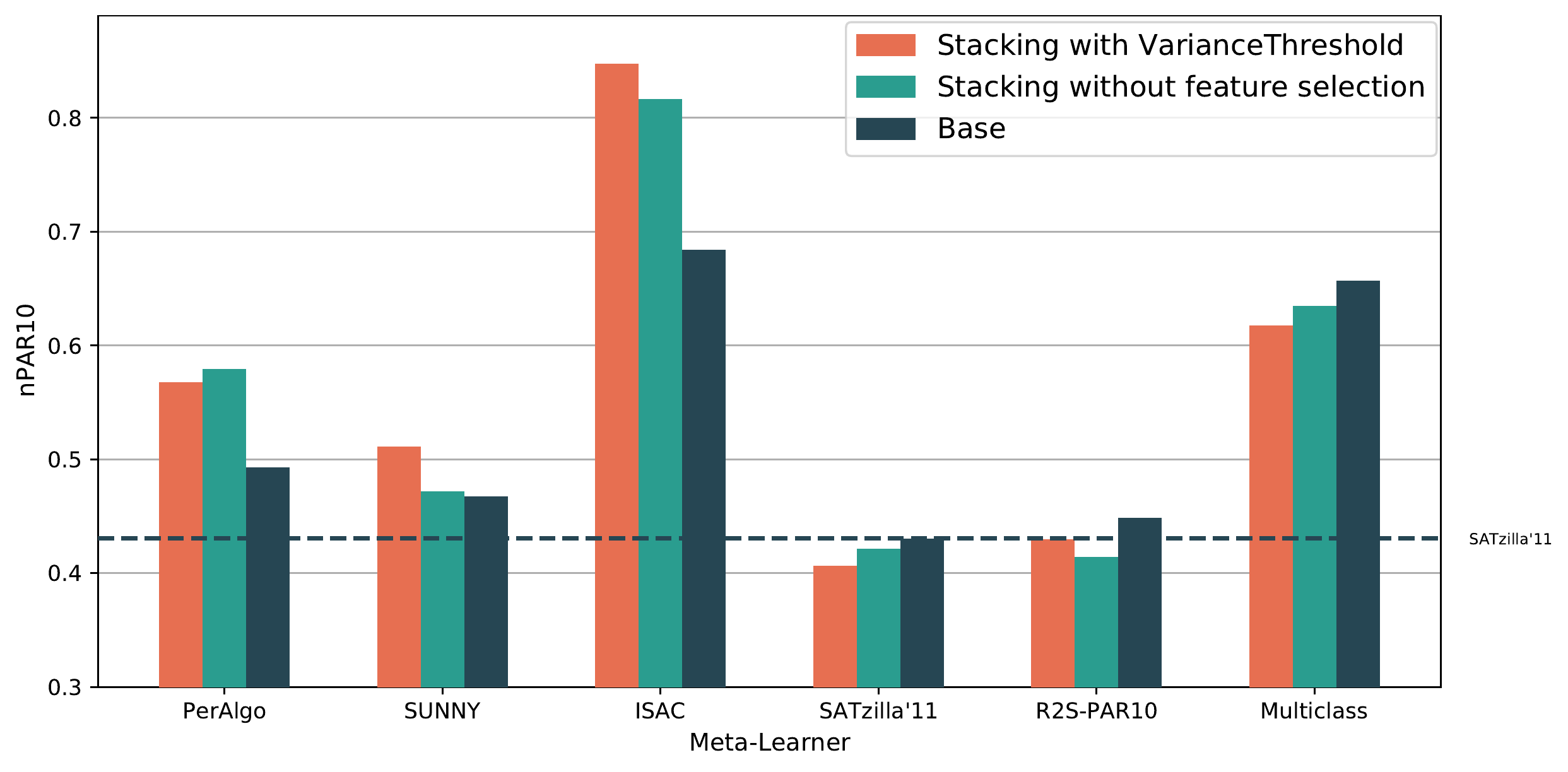}
    \caption{This figure shows the average \textit{nPAR10} performance of stacking variants where $h_\mathit{agg}$, i.e. the meta-learner, is instantiated through different algorithm selectors with and without a variance threshold feature selection approaches.}
    \label{fig:stacking_bar_chart}
\end{figure}

\autoref{fig:stacking_bar_chart} shows the average \textit{nPAR10} performance of stacking variants, where the meta-learner $h_\mathit{agg}$ is instantiated through different algorithm selectors with and without a variance threshold feature selection approach. Each variant uses all base algorithm selectors to generate additional features. The variance threshold method selects all features with a variance larger than a given threshold, which was set to $0.16$ for these experiments. The dotted line indicates the average performance of the SBAS. 

\begin{figure}[t]
    \centering
    \includegraphics[width=\textwidth]{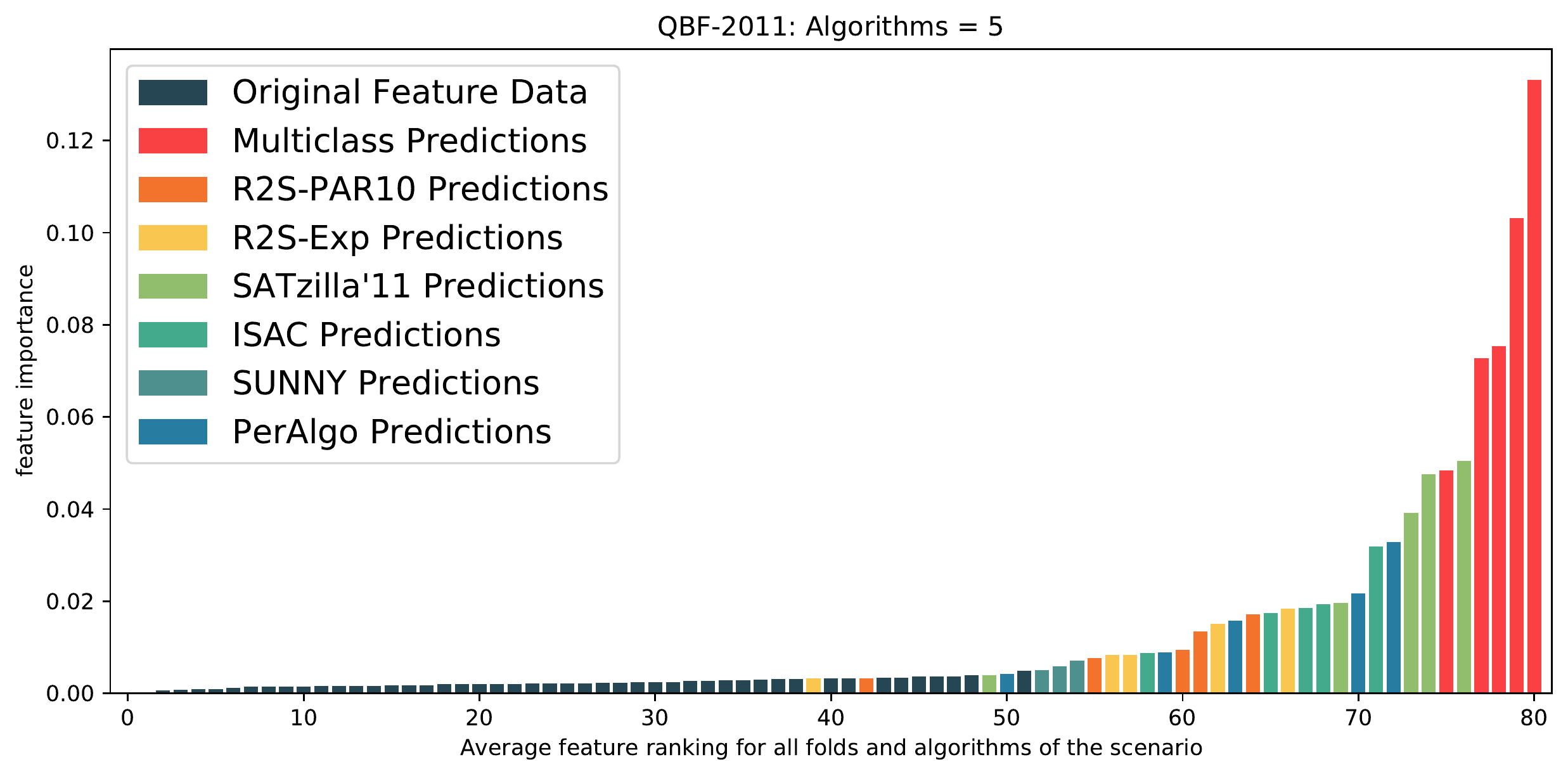}
    \caption{This figure portrays a ranking over the features w.r.t.\ their feature importance values extracted from the multi-class classification meta-learner (instantiated with a one-vs-all decomposition equipped with a random forest classifier) for the QBF-2011 scenario.}
    \label{fig:stacking_feature_importance_values}
\end{figure}

Firstly, we would like to note that no general recommendation on the use of feature selection can be made, as the effect seems to depend very much on the meta-learner.
However, while the majority of stacking ensemble variants do not improve over the best algorithm selector, variants deploying SATzilla'11 and R2S-PAR10 as a meta-learner can slightly improve in performance. We find this quite disappointing, because the additional features provided to the meta-learner seem to carry valuable information. This is confirmed by the feature importance analysis portrayed in \autoref{fig:stacking_feature_importance_values}. It shows a ranking over the features w.r.t.\ their feature importance values extracted from the multi-class classification meta-learner (instantiated with a random forest classifier) for the QBF-2011 scenario. Clearly, the additional features in the form of the predictions of the ensemble members carry the biggest part of the information contained in the data.

\subsubsection{Overall Comparison}
\autoref{tab:overall_results} displays \textit{nPAR10} values of a subset of all evaluated ensemble variants and all base algorithm selectors broken down by the different scenarios. The best result for each scenario is marked in bold, and a line above a result of an ensemble approach indicates that it is better than the result of the best base algorithm selector on the corresponding scenario.

Overall, ensembles of algorithm selectors achieve a performance superior to single algorithm selectors. There is only a single scenario (ASP-POTASSCO) for which none of the selected ensemble variants was able to improve over the base algorithm selector, performing best on that particular scenario, and another scenario where a performance on par was achieved (MAXSAT15-PMS-INDU). For all other scenarios, at least one of the ensemble variants achieved a new state-of-the-art performance. While some of these improvements are rather small (CSP-MNZ-201, where an improvement from $0.11$ to $0.10$ is recorded), there are also various scenarios with a $>1.5$ fold improvement (e.g., CSP-Minizinc-Time-2016, SAT03\_16\_INDU, QBF-2011). This is especially remarkable as only very few improvements have been made in the last two years.

In terms of mean, median, and average rank performance across all scenarios, both voting ensemble variants achieve the best result and improve over the previous state of the art by more than $40\%$ (median performance). Thus, they demonstrate a very robust performance across all scenarios. The voting ensemble with a Borda aggregation, the bagging ensemble with the PerAlgo base selector and a Borda aggregation, and the boosting ensemble with the PerAlgo base selector and a weighted majority aggregation all consistently outperform the best single algorithm selector on $6$ of $13$ scenarios, and thus achieve an impressive performance.

\begin{table}[t]
    \centering
    \caption{\textit{nPAR10} values of the best ensemble variants and all base algorithm selectors broken down by the different scenarios. The best result for each scenario is marked in bold and a line above a result indicates beating all base algorithm selectors.}
    \resizebox{\textwidth}{!}{
    \begin{tabular}{l | r r | r r | r r | r r || r r r r r r r }
         \toprule
         \textbf{Ensemble} & 
            \multicolumn{2}{c|}{Voting} & 
            \multicolumn{2}{c|}{Bagging} & 
            \multicolumn{2}{c|}{Stacking} & 
            \multicolumn{2}{c||}{Boosting} & \\ \midrule
         \textbf{Aggregation} & 
            \rotatebox{90}{wmaj} & \rotatebox{90}{borda} & 
            \rotatebox{90}{wmaj} & \rotatebox{90}{borda} & 
            \rotatebox{90}{R2S-Exp} & \rotatebox{90}{SATzilla'11 (VT)} &
            \rotatebox{90}{wmaj} & \rotatebox{90}{wmaj} &\\ \midrule
         \textbf{Base selector} & 
            \rotatebox{90}{all} & \rotatebox{90}{all} & 
            \rotatebox{90}{SUNNY} & \rotatebox{90}{PerAlgo} & 
            \rotatebox{90}{all} & \rotatebox{90}{all} & 
            \rotatebox{90}{Multiclass} & \rotatebox{90}{PerAlgo} &
            \rotatebox{90}{R2S-Exp} & 
            \rotatebox{90}{R2S-PAR10} & 
            \rotatebox{90}{ISAC} & 
            \rotatebox{90}{Multiclass} & 
            \rotatebox{90}{PerAlgo} & 
            \rotatebox{90}{SATzilla'11} & 
            \rotatebox{90}{SUNNY}\\ 
        \midrule
        \textbf{Scenario} \\
        \midrule
         ASP-POTASSCO & 0.26 & 0.24 & 0.21 & 0.29 & 0.31 & 0.31 & 0.73 & 0.44 & 0.30 & 0.34 & 0.64 & 0.67 & 0.34 & 0.45 & \textbf{0.17} \\
         BNSL-2016 & $\overline{0.16}$ & $\overline{0.17}$ & 0.25 & \textbf{0.15} & $\overline{0.16}$ & 0.18 & 0.70 & 0.32 & 0.20 & 0.22 & 0.84 & 0.31 & 0.20 & 0.18 & 0.25 \\
         CPMP-2015 & 0.81 & 0.87 & 0.83 & 0.82 & 0.88 & $\overline{0.76}$ & $\overline{0.69}$ & $\overline{\mathbf{0.51}}$ & 0.97 & 0.81 & 0.98 & 0.94 & 0.90 & 0.81 & 1.05 \\
         CSP-2010 & 0.24 & 0.24 & 0.33 & $\overline{\mathbf{0.23}}$ & $\overline{\mathbf{0.23}}$ & 0.24 & 0.63 & 0.43 & 0.26 & 0.26 & 0.38 & 0.78 & 0.36 & 0.24 & 0.40 \\
         CSP-MZN-2013 & $\overline{\mathbf{0.10}}$ & 0.11 & 0.12 & $\overline{\mathbf{0.10}}$ & 0.14 & 0.20 & 0.57 & 0.39 & 0.11 & 0.11 & 0.34 & 0.31 & 0.13 & 0.22 & 0.13 \\
         CSP-Minizinc-Time-2016 & $\overline{\mathbf{0.21}}$ & $\overline{0.31}$ & 0.51 & 0.51 & 0.46 & $\overline{0.40}$ & 0.52 & $\overline{0.39}$ & 0.46 & 0.46 & 0.70 & 0.61 & 0.61 & 0.41 & 0.52 \\
         GLUHACK-18 & 0.44 & 0.44 & 0.47 & 0.49 & 0.45 & 0.43 & 0.49 & $\overline{\mathbf{0.38}}$ & 0.47 & 0.50 & 0.60 & 0.39 & 0.44 & 0.41 & 0.52 \\
         MAXSAT12-PMS & 0.27 & 0.27 & $\overline{\mathbf{0.17}}$ & $\overline{0.21}$ & 0.28 & 0.33 & 0.45 & 0.34 & 0.27 & 0.29 & 0.55 & 0.37 & 0.33 & 0.24 & 0.28 \\
         MAXSAT15-PMS-INDU & 0.36 & \textbf{0.24} & 0.31 & 0.40 & 0.34 & 0.30 & 0.44 & 0.34 & 0.39 & 0.46 & 1.00 & 1.24 & 0.58 & 0.43 & \textbf{0.24} \\
         QBF-2011 & 0.18 & 0.17 & 0.16 & $\overline{\mathbf{0.10}}$ & 0.16 & $\overline{0.14}$ & 0.42 & 0.33 & 0.19 & 0.19 & 0.33 & 0.33 & 0.20 & 0.16 & 0.22 \\
         SAT03-16\_INDU & 0.73 & $\overline{0.71}$ & $\overline{0.70}$ & 0.75 & $\overline{0.66}$ & 0.80 & $\overline{0.43}$ & $\overline{\mathbf{0.38}}$ & 0.72 & 0.76 & 0.94 & 0.99 & 0.89 & 0.84 & 0.85 \\
         SAT12-INDU & 0.61 & $\overline{0.58}$ & 0.71 & 0.73 & 0.73 & $\overline{0.59}$ & $\overline{0.45}$ & $\overline{\mathbf{0.42}}$ & 0.73 & 0.75 & 0.97 & 0.94 & 0.81 & 0.61 & 0.81 \\
         SAT18-EXP & $\overline{0.47}$ & $\overline{0.52}$ & $\overline{0.57}$ & $\overline{\mathbf{0.38}}$ & $\overline{0.52}$ & $\overline{0.59}$ & $\overline{0.47}$ & $\overline{0.42}$ & 0.60 & 0.67 & 0.61 & 0.65 & 0.62 & 0.60 & 0.63 \\
         \midrule \midrule
         Mean & $\overline{\mathbf{0.37}}$ & $\overline{\mathbf{0.37}}$ & $\overline{0.41}$ & $\overline{0.40}$ & $\overline{0.41}$ & $\overline{0.41}$ & 0.54 & $\overline{0.39}$ & 0.44 & 0.45 & 0.68 & 0.66 & 0.49 & 0.43 & 0.47 \\
         Median & $\overline{\mathbf{0.27}}$ & $\overline{\mathbf{0.27}}$ & $\overline{0.33}$ & $\overline{0.38}$ & $\overline{0.34}$ & $\overline{0.33}$ & 0.49 & 0.39 & 0.39 & 0.46 & 0.64 & 0.65 & 0.44 & 0.41 & 0.40 \\
         Agv. Rank & $\overline{\mathbf{4.38}}$ & $\overline{\mathbf{4.38}}$ & $\overline{6.31}$ & $\overline{5.08}$ & $\overline{6.08}$ & $\overline{5.92}$ & 10.08 & $\overline{6.85}$ & 7.62 & 9.46 & 13.69 & 12.69 & 10.46 & 7.00 & 9.92 \\
\bottomrule
    \end{tabular}
    }
    \label{tab:overall_results}
\end{table}

\subsection{Discussion of Results: Is Meta Learning Harder Than Learning?}
Recall our taxonomy of the approaches presented in \autoref{fig:approach_relation}, regarding which kind of mapping they model, how this mapping is constructed, and how the required aggregation function is obtained. 
Drawing an overall conclusion from the results presented in this work, we cautiously conclude that the presumably simpler problem of learning a mapping (\ref{eq:meta_learn_as}) from the instances to the set of algorithm selectors yields worse results than solving the presumably more complicated problem of finding both a mapping from instances to a set of selectors and a corresponding aggregation function. While we observed remarkable performance improvements for all ensemble approaches, the meta learning approach could essentially achieve no improvement. 

As a possible reason, note that the meta learning approach heavily relies on the instance features, which are required for learning on the meta level. On the contrary, ensembles of algorithm selectors do not use these features on the meta level directly  (except for stacking), but only aggregate the predictions of multiple selectors. Thus, we speculate that the information contained in the features does not allow for an improvement in performance through moving to the meta level, while the predictions of the selectors do carry enough information to do so. This hypothesis is corroborated by the feature analysis conducted as part of the experiments around stacking (cf.\ \autoref{fig:stacking_feature_importance_values}), which indicate that much more information is present in the predictions of the base selectors than in the original instance features. We attribute stacking's ability to perform successful learning on the meta level (aggregation) to the same reason. While stacking was able to achieve improvements, the arguably most simple ensemble approach in the form of voting, which involves no learning on the meta level at all, achieved by far the best results.
Overall, learning on the meta level appears to be a very hard problem.

\section{Related Work}\label{sec:related-work}
In the following, we give an overview of the most related work regarding the use of ensemble methods in algorithm selection. As mentioned earlier, this work is surprisingly sparse. For a general overview of work on algorithm selection, we refer to \cite{kerschkeHNT19_as_survey}.

We presented a preliminary version of the meta AS problem in a preprint \cite{metaas}, which aimed at constructing a more effective algorithm selector by leveraging multiple existing selectors. The idea presented there is identical to the idea presented here in \autoref{sec:meta-learning_ass}. In this work, we define the problem in a more general fashion, present a framework for solving this problem and show several instantiations of this framework. Accordingly, the work presented in the preprint is subsumed by this work.

In algorithm selection, it is normally assumed that the set of algorithms $\algorithms$ to choose from is predefined, although the composition of this set can have an influence on the selectors. Therefore, \cite{kordikCF18} propose to not simply use all available algorithms as a basis to choose from, but to employ ensemble techniques in order to construct algorithms constituting this set. Thus, \cite{kordikCF18} build ensembles on the level of algorithms, whereas we ensemble on the level of selectors with the goal to create a better combined algorithm selector.

Last but not least, and perhaps indeed most related, both \cite{malone2017asl} and \cite{kotthoff12} suggest a stacking approach: First, a regression model is learned per algorithm to estimate the performance on a given instance, and second, the estimated performances are used as input for a multi-class classification model that eventually selects the algorithm. While \cite{kotthoff12} only uses the outputs of the performance estimators as input of the meta-learner, \cite{malone2017asl} use these in addition to the original features. Moreover, \cite{malone2017asl} suggest to also include uncertainty information obtained from the performance estimators as input for the meta-learner. Both variants are very specific instantiations of the general idea presented in this paper, using stacking as an ensemble technique and a specific selector as a base algorithm selector. While the approach presented by \cite{malone2017asl} resulted in the last stop in the open algorithm selection competition of 2017 \cite{lindauerRK19_as_competitions}, \cite{kotthoff12} considered a setting, where the goal was to select the best machine learning algorithm for a dataset. He showed that stacking a classifier on top of the pure performance estimation does yield indeed an improvement in most cases over choosing the algorithm based on the performance estimates only.

\section{Conclusion}\label{sec:conclusion}

In this paper, we revisited the problem of algorithm selection from a meta perspective. We defined the problem of meta algorithm selection and proposed a general methodological framework for this problem. Moreover, we considered several concrete learning methods as instantiations of this framework and compared them conceptually and empirically. In an extensive experimental study on an established benchmark for algorithm selection, we have shown that the meta algorithm selection problem can be solved efficiently, and that solutions can provide remarkable improvements in performance, often significantly better than the hitherto state of the art. Finally, we set the results into a broader context, concluding that learning algorithm selector selectors seems to be harder and less promising than defining them through well-established concepts from ensemble learning.

In future work, more effort should be invested in understanding why learning algorithm selector selectors appears to be a hard problem, while manually defined algorithm selection ensembles can achieve good performance. In particular, investigations of this phenomenon on a theoretical level would be of interest.
Another possible direction for future work might be to focus more on learning instance-specific aggregation functions \cite{melnikov2016learning} to be used inside the ensembles, because this would allow one to leverage the information of which algorithm did indeed perform best on a given instance, instead of using an a priori fixed aggregation function. As seen with stacking, this works at least in principle.
Yet another direction for future work is to adapt the idea of ensembles to the field of algorithm scheduling, where the recommendation target is no longer a single algorithm, but a complete algorithm schedule. One of the main challenges here is the aggregation of schedules.

\section*{Acknowledgements}
This work was partially supported by the German Research Foundation (DFG) within the Collaborative Research Center ``On-The-Fly Computing'' (SFB 901/3 project no.\ 160364472) and the German Federal Ministry of Education and Research (ITS.ML project no.\ 01IS18041D). The authors gratefully acknowledge support of this project through computing time provided by the Paderborn Center for Parallel Computing (PC$^2$).

%
%

\bibliographystyle{splncs04}
\bibliography{bibliography}   

\end{document}

%% file: figures/scenarios.tex
\definecolor{winner}{RGB}{240,240,240}

\begin{tabular}{llllllllllllll}
                          & \multicolumn{1}{c}{\rotatebox{90}{ASP-POTASSCO}} & \multicolumn{1}{c}{\rotatebox{90}{BNSL-2016}} & \multicolumn{1}{c}{\rotatebox{90}{CPMP-2015}} & \multicolumn{1}{c}{\rotatebox{90}{CSP-2010}} & \multicolumn{1}{c}{\rotatebox{90}{CSP-MZN-2013}} & \multicolumn{1}{c}{\rotatebox{90}{CSP-Minizinc-Time-2016  }} & \multicolumn{1}{c}{\rotatebox{90}{GLUHACK-18}} & \multicolumn{1}{c}{\rotatebox{90}{MAXSAT12-PMS}} & \multicolumn{1}{c}{\rotatebox{90}{MAXSAT15-PMS-INDU}} & \multicolumn{1}{c}{\rotatebox{90}{QBF-2011}} & \multicolumn{1}{c}{\rotatebox{90}{SAT03-16\_INDU}} & \multicolumn{1}{c}{\rotatebox{90}{SAT12-INDU}} & \multicolumn{1}{c}{\rotatebox{90}{SAT18-EXP}} \\ \hline
\multicolumn{1}{l|}{\#I}          & \multicolumn{1}{l|}{1294}        & \multicolumn{1}{l|}{1179}     & \multicolumn{1}{l|}{527}      & \multicolumn{1}{l|}{2024}    & \multicolumn{1}{l|}{4642}        & \multicolumn{1}{l|}{100}                   & \multicolumn{1}{l|}{353}          & \multicolumn{1}{l|}{876}         & \multicolumn{1}{l|}{601}              & \multicolumn{1}{l|}{1368}    & \multicolumn{1}{l|}{2000}          & \multicolumn{1}{l|}{1167}      & \multicolumn{1}{l|}{353}                              \\ \hline
\multicolumn{1}{l|}{\#U} & \multicolumn{1}{l|}{82}          & \multicolumn{1}{l|}{0}        & \multicolumn{1}{l|}{0}        & \multicolumn{1}{l|}{253}     & \multicolumn{1}{l|}{944}         & \multicolumn{1}{l|}{17}                    & \multicolumn{1}{l|}{116}          & \multicolumn{1}{l|}{129}         & \multicolumn{1}{l|}{44}               & \multicolumn{1}{l|}{314}     & \multicolumn{1}{l|}{269}           & \multicolumn{1}{l|}{209}       &   \multicolumn{1}{l|}{67}    \\ \hline
\multicolumn{1}{l|}{\#A}         & \multicolumn{1}{l|}{11}          & \multicolumn{1}{l|}{8}        & \multicolumn{1}{l|}{4}        & \multicolumn{1}{l|}{2}       & \multicolumn{1}{l|}{11}          & \multicolumn{1}{l|}{20}                    & \multicolumn{1}{l|}{8}          & \multicolumn{1}{l|}{6}           & \multicolumn{1}{l|}{29}               & \multicolumn{1}{l|}{5}       & \multicolumn{1}{l|}{10}            & \multicolumn{1}{l|}{31}        & \multicolumn{1}{l|}{37}                              \\ \hline
\multicolumn{1}{l|}{\#F}           & \multicolumn{1}{l|}{138}         & \multicolumn{1}{l|}{86}       & \multicolumn{1}{l|}{22}       & \multicolumn{1}{l|}{86}      & \multicolumn{1}{l|}{155}         & \multicolumn{1}{l|}{95}                    & \multicolumn{1}{l|}{50}          & \multicolumn{1}{l|}{37}          & \multicolumn{1}{l|}{37}               & \multicolumn{1}{l|}{46}      & \multicolumn{1}{l|}{483}           & \multicolumn{1}{l|}{115}       &  \multicolumn{1}{l|}{50}                              \\ \hline
\multicolumn{1}{l|}{C}             & \multicolumn{1}{l|}{600}         & \multicolumn{1}{l|}{7200}     & \multicolumn{1}{l|}{3600}     & \multicolumn{1}{l|}{5000}    & \multicolumn{1}{l|}{1800}        & \multicolumn{1}{l|}{1200}                  & \multicolumn{1}{l|}{5000}          & \multicolumn{1}{l|}{2100}        & \multicolumn{1}{l|}{2100}             & \multicolumn{1}{l|}{3600}    & \multicolumn{1}{l|}{5000}          & \multicolumn{1}{l|}{1200}      & \multicolumn{1}{l|}{5000}    \\
\hline                          
\end{tabular}

%% file: tables/n_par_10_all_normalized_by_level_0.tex
\begin{tabular}{l||rrrrrrr|rrrrrrr}
\toprule
\multicolumn{1}{c||}{Level} & \multicolumn{7}{c}{Algorithm Selectors} & \multicolumn{7}{|c}{Algorithm Selector Selectors (Meta)} \\
\toprule
\multicolumn{1}{r||}{\rotatebox{90}{Approach}} &  \multicolumn{1}{c}{\rotatebox{90}{R2SExp}} &  \multicolumn{1}{c}{\rotatebox{90}{R2SPAR10}} &   \rotatebox{90}{ISAC} &    \multicolumn{1}{c}{\rotatebox{90}{MCC}} & \multicolumn{1}{c}{\rotatebox{90}{PAReg}} &  \multicolumn{1}{c}{\rotatebox{90}{SATzilla'11}} &  \rotatebox{90}{SUNNY} &  \multicolumn{1}{c}{\rotatebox{90}{R2SExp}} &  \multicolumn{1}{c}{\rotatebox{90}{R2SPAR10}} &  \multicolumn{1}{c}{\rotatebox{90}{ISAC}} &  \multicolumn{1}{c}{\rotatebox{90}{MCC}} &  \multicolumn{1}{c}{\rotatebox{90}{PAReg}} &  \multicolumn{1}{c}{\rotatebox{90}{SATzilla'11}} &  \multicolumn{1}{c}{\rotatebox{90}{SUNNY}} \\
Scenario         &         &           &        &        &              &        &            &              &          &         &           &                 &           \\
\midrule

ASP-POTASSCO & 0.30 & 0.32 & 0.63 & 0.64 & 0.34 & 0.47 & \textbf{0.17} & 0.26 (6/1) & 0.21 (6/1) & 0.26 (6/1) & 0.37 (3/4) & 0.31 (5/2) & 0.29 (6/1) & 0.26 (6/1)\\
BNSL-2016 & \textbf{0.18} & 0.21 & 0.84 & 0.31 & \textbf{0.18} & \textbf{0.18} & 0.25 & 0.23 (3/4) & 0.21 (4/3) & 0.19 (4/3) & 0.24 (3/4) & 0.23 (3/4) & 0.28 (2/5) & 0.27 (2/5)\\
CPMP-2015 & 0.76 & \textbf{0.69} & 0.90 & 0.85 & 0.78 & 0.70 & 0.94 & 0.78 (4/3) & 0.78 (4/3) & 0.89 (2/5) & 0.81 (3/4) & 0.77 (4/3) & 0.81 (3/4) & 0.89 (2/5)\\
CSP-2010 & 0.13 & 0.15 & 0.31 & 0.80 & 0.25 & 0.13 & 0.34 & \textbf{0.04} (7/0) & 0.10 (7/0) & 0.19 (4/3) & 0.13 (7/0) & 0.46 (1/6) & 0.18 (4/3) & 0.09 (7/0)\\
CSP-MZN-2013 & \textbf{0.11} & \textbf{0.11} & 0.35 & 0.31 & 0.13 & 0.21 & 0.13 & \textbf{0.11} (7/0) & \textbf{0.11} (7/0) & 0.13 (5/2) & 0.15 (3/4) & 0.13 (5/2) & 0.19 (3/4) & 0.14 (3/4)\\
CSP-Minizinc-Time-2016 & 0.43 & \textbf{0.27} & 0.83 & 0.36 & 0.67 & 0.34 & 0.37 & 0.51 (2/5) & 0.51 (2/5) & 0.76 (1/6) & 0.60 (2/5) & 0.67 (2/5) & 0.35 (5/2) & 0.51 (2/5)\\
GLUHACK-18 & 0.43 & 0.46 & 0.69 & 0.41 & 0.46 & 0.42 & 0.51 & \textbf{0.40} (7/0) & 0.45 (4/3) & 0.41 (7/0) & 0.49 (2/5) & 0.57 (1/6) & 0.47 (2/5) & 0.46 (4/3)\\
MAXSAT12-PMS & 0.22 & 0.23 & 0.47 & 0.40 & 0.28 & 0.24 & 0.29 & 0.25 (4/3) & 0.25 (4/3) & 0.20 (7/0) & \textbf{0.19} (7/0) & 0.32 (2/5) & 0.20 (7/0) & 0.21 (7/0)\\
MAXSAT15-PMS-INDU & 0.34 & 0.44 & 0.89 & 1.06 & 0.55 & 0.39 & \textbf{0.24} & 0.36 (5/2) & 0.57 (2/5) & 0.33 (6/1) & 0.39 (5/2) & 0.40 (4/3) & 0.51 (3/4) & 0.26 (6/1)\\
QBF-2011 & 0.21 & 0.20 & 0.36 & 0.35 & 0.18 & \textbf{0.15} & 0.22 & 0.22 (3/4) & 0.22 (3/4) & 0.21 (4/3) & 0.21 (4/3) & 0.29 (2/5) & 0.23 (2/5) & 0.26 (2/5)\\
SAT03-16\_INDU & \textbf{0.71} & 0.76 & 0.98 & 0.99 & 0.77 & 0.82 & 0.82 & 0.92 (2/5) & 0.90 (2/5) & 0.81 (4/3) & 0.79 (4/3) & 0.81 (4/3) & 0.84 (2/5) & 0.86 (2/5)\\
SAT12-INDU & 0.70 & 0.73 & 1.02 & 0.94 & 0.79 & \textbf{0.59} & 0.78 & 0.62 (6/1) & 0.63 (6/1) & 0.75 (4/3) & 0.73 (5/2) & 0.65 (6/1) & 0.65 (6/1) & 0.66 (6/1)\\
SAT18-EXP & 0.61 & 0.68 & 0.62 & 0.65 & 0.64 & 0.59 & 0.63 & 0.66 (1/6) & 0.67 (1/6) & 0.61 (6/1) & 0.58 (7/0) & 0.61 (6/1) & \textbf{0.54} (7/0) & 0.59 (7/0)\\
\bottomrule
\end{tabular}

%% file: main_arxiv.bbl
\begin{thebibliography}{10}
\providecommand{\url}[1]{\texttt{#1}}
\providecommand{\urlprefix}{URL }
\providecommand{\doi}[1]{https://doi.org/#1}

\bibitem{sunny_amadiniGM14}
Amadini, R., Gabbrielli, M., Mauro, J.: {SUNNY:} a lazy portfolio approach for
  constraint solving. Theory Pract. Log. Program.  \textbf{14}(4-5) (2014)

\bibitem{bischlKKLMFHHLT16}
Bischl, B., Kerschke, P., Kotthoff, L., Lindauer, M., Malitsky, Y.,
  Fr{\'{e}}chette, A., Hoos, H.H., Hutter, F., Leyton{-}Brown, K., Tierney, K.,
  Vanschoren, J.: Aslib: {A} benchmark library for algorithm selection. Artif.
  Intell.  \textbf{237},  41--58 (2016)

\bibitem{borda1784memoire_borda_count}
Borda, J.d.: M{\'e}moire sur les {\'e}lections au scrutin. Histoire de
  l'Academie Royale des Sciences pour 1781 (Paris, 1784)  (1784)

\bibitem{breiman96b_bagging}
Breiman, L.: Bagging predictors. Mach. Learn.  \textbf{24}(2),  123--140 (1996)

\bibitem{copp_ob06}
Coppersmith, D., Fleischer, L., Rudra, A.: Ordering by weighted number of wins
  gives a good ranking for weighted tournaments. In: ACM-SIAM Symposium on
  Discrete Algorithms (SODA). pp. 776--782 (2006)

\bibitem{dietterich00_ensembles}
Dietterich, T.G.: Ensemble methods in machine learning. In: Multiple Classifier
  Systems, First International Workshop, {MCS} 2000, Cagliari, Italy, June
  21-23, 2000, Proceedings. pp. 1--15 (2000)

\bibitem{drucker1997improving}
Drucker, H.: Improving regressors using boosting techniques. In: ICML. vol.~97,
  pp. 107--115. Citeseer (1997)

\bibitem{dworkKNS01_rank_aggregation}
Dwork, C., Kumar, R., Naor, M., Sivakumar, D.: Rank aggregation methods for the
  web. In: Proceedings of the Tenth International World Wide Web Conference,
  {WWW} 10, Hong Kong, China, May 1-5, 2001. pp. 613--622 (2001)

\bibitem{frossyniotis2004clustering}
Frossyniotis, D., Likas, A., Stafylopatis, A.: A clustering method based on
  boosting. Pattern Recognition Letters  \textbf{25}(6),  641--654 (2004)

\bibitem{garcia2009boosting}
Garc{\'\i}a-Pedrajas, N., Ortiz-Boyer, D.: Boosting k-nearest neighbor
  classifier by means of input space projection. Expert Systems with
  Applications  \textbf{36}(7),  10570--10582 (2009)

\bibitem{gomesSC97_heavy_tails}
Gomes, C.P., Selman, B., Crato, N.: Heavy-tailed distributions in combinatorial
  search. In: Principles and Practice of Constraint Programming - CP97, Third
  International Conference, Linz, Austria, October 29 - November 1, 1997,
  Proceedings. pp. 121--135 (1997)

\bibitem{guyonE03_feature_selection}
Guyon, I., Elisseeff, A.: An introduction to variable and feature selection. J.
  Mach. Learn. Res.  \textbf{3},  1157--1182 (2003)

\bibitem{crr_hanselle2020}
Hanselle, J., Tornede, A., Wever, M., Hüllermeier, E.: Hybrid ranking and
  regression for algorithm selection. In: KI 2020: Advances in Artificial
  Intelligence (2020)

\bibitem{pakdd_hanselle2021algorithm}
Hanselle, J., Tornede, A., Wever, M., H{\"u}llermeier, E.: Algorithm selection
  as superset learning: Constructing algorithm selectors from imprecise
  performance data. In: The 25th Pacific-Asia Conference on Knowledge Discovery
  and Data Mining (PAKDD-2021), May 11-14, 2021, Delhi, India. (2021)

\bibitem{hastie2009multi}
Hastie, T., Rosset, S., Zhu, J., Zou, H.: Multi-class adaboost. Statistics and
  its Interface  \textbf{2}(3),  349--360 (2009)

\bibitem{superset_learning_HULLERMEIER20141519}
Hüllermeier, E.: Learning from imprecise and fuzzy observations: Data
  disambiguation through generalized loss minimization. International Journal
  of Approximate Reasoning  \textbf{55}(7),  1519 -- 1534 (2014), special
  issue: Harnessing the information contained in low-quality data sources

\bibitem{isac_kadiogluMST10}
Kadioglu, S., Malitsky, Y., Sellmann, M., Tierney, K.: {ISAC} -
  instance-specific algorithm configuration. In: {ECAI} (2010)

\bibitem{kerschkeHNT19_as_survey}
Kerschke, P., Hoos, H.H., Neumann, F., Trautmann, H.: Automated algorithm
  selection: Survey and perspectives. Evol. Comput.  \textbf{27}(1),  3--45
  (2019)

\bibitem{kordikCF18}
Kord{\'{\i}}k, P., Cern{\'{y}}, J., Fr{\'{y}}da, T.: Discovering predictive
  ensembles for transfer learning and meta-learning. Mach. Learn.
  \textbf{107}(1),  177--207 (2018)

\bibitem{kotthoff12}
Kotthoff, L.: Hybrid regression-classification models for algorithm selection.
  In: {ECAI} 2012 - 20th European Conference on Artificial Intelligence. (2012)

\bibitem{lindauerRK19_as_competitions}
Lindauer, M., van Rijn, J.N., Kotthoff, L.: The algorithm selection
  competitions 2015 and 2017. Artif. Intell.  \textbf{272},  86--100 (2019)

\bibitem{lobjois1998branch}
Lobjois, L., Lema{\^\i}tre, M., et~al.: Branch and bound algorithm selection by
  performance prediction. In: AAAI/IAAI. pp. 353--358 (1998)

\bibitem{malone2017asl}
Malone, B., Kangas, K., J{\"a}rvisalo, M., Koivisto, M., Myllym{\"a}ki, P.:
  as-asl: Algorithm selection with auto-sklearn. In: Open Algorithm Selection
  Challenge 2017. pp. 19--22. PMLR (2017)

\bibitem{melnikov2016learning}
Melnikov, V., H{\"u}llermeier, E.: Learning to aggregate using uninorms. In:
  Joint European Conference on Machine Learning and Knowledge Discovery in
  Databases. pp. 756--771. Springer (2016)

\bibitem{DBLP:conf/ictai/PiheraM14}
Pihera, J., Musliu, N.: Application of machine learning to algorithm selection
  for {TSP}. In: 26th {IEEE} International Conference on Tools with Artificial
  Intelligence, {ICTAI} 2014, Limassol, Cyprus, November 10-12, 2014. pp.
  47--54. {IEEE} Computer Society (2014)

\bibitem{rice1976algorithm}
Rice, J.R.: The algorithm selection problem. In: Advances in computers,
  vol.~15, pp. 65--118. Elsevier (1976)

\bibitem{saari2000mathematics_borda_extension}
Saari, D.G.: The mathematics of voting: Democratic symmetry. Economist
  \textbf{83} (2000)

\bibitem{schapire1990strength_boosting}
Schapire, R.E.: The strength of weak learnability. Machine learning
  \textbf{5}(2),  197--227 (1990)

\bibitem{extreme_algorithm_selection_tornedeWH20}
Tornede, A., Wever, M., H{\"{u}}llermeier, E.: Extreme algorithm selection with
  dyadic feature representation. In: Discovery Science (2020)

\bibitem{metaas}
Tornede, A., Wever, M., H{\"u}llermeier, E.: Towards meta-algorithm selection.
  In: {Workshop on Meta-Learning (MetaLearn 2020) @ NeurIPS 2020} (2020)

\bibitem{tornede2019algorithm}
Tornede, A., Wever, M., Hüllermeier, E.: Algorithm selection as
  recommendation: From collaborative filtering to dyad ranking. In: CI
  Workshop, Dortmund (2019)

\bibitem{tornede20_run2survive}
Tornede, A., Wever, M., Werner, S., Mohr, F., H{\"u}llermeier, E.: Run2survive:
  A decision-theoretic approach to algorithm selection based on survival
  analysis. In: ACML (2020)

\bibitem{meta_learning_survey}
Vanschoren, J.: Meta-learning: {A} survey. CoRR  \textbf{abs/1810.03548} (2018)

\bibitem{tpami}
{Wever}, M., {Tornede}, A., {Mohr}, F., {H{\"u}llermeier}, E.: Automl for
  multi-label classification: Overview and empirical evaluation. IEEE
  Transactions on Pattern Analysis and Machine Intelligence pp.~1--1 (2021)

\bibitem{no_free_lunch_wolpert1997no}
Wolpert, D.H., Macready, W.G., et~al.: No free lunch theorems for optimization.
  Evol. Comput.  \textbf{1}(1) (1997)

\bibitem{wolpert92_stacking}
Wolpert, D.H.: Stacked generalization. Neural Networks  \textbf{5}(2),
  241--259 (1992)

\bibitem{satzilla11_xu2011hydra}
Xu, L., Hutter, F., Hoos, H., Leyton-Brown, K.: Hydra-mip: Automated algorithm
  configuration and selection for mixed integer programming. In: {RCRA workshop
  @ IJCAI} (2011)

\bibitem{satzilla07_xu2007}
Xu, L., Hutter, F., Hoos, H.H., Leyton-Brown, K.: Satzilla-07: the design and
  analysis of an algorithm portfolio for sat. In: {CP}. Springer (2007)

\end{thebibliography}
